\documentclass[11pt]{article}

\usepackage[final]{acl}
\usepackage{times}
\usepackage{latexsym}
\usepackage{inconsolata}

\usepackage[utf8]{inputenc}
\usepackage[T1]{fontenc}
\usepackage{url}
\usepackage{booktabs}
\usepackage{amsfonts,amsmath,amssymb,amsthm}
\usepackage{nicefrac}
\usepackage{microtype}
\usepackage{xcolor}
\usepackage{colortbl}
\usepackage{graphicx}
\usepackage{multirow}
\usepackage{algorithm}
\usepackage{algpseudocode}
\usepackage[inline]{enumitem}
\usepackage{pifont}
\usepackage{dsfont}
\usepackage[most]{tcolorbox}
\usepackage{soul}
\usepackage{etoc}

\DeclareMathOperator*{\argmax}{arg\,max}
\newcommand{\Tmax}{T_{\max}}
\newcommand{\eos}{\varepsilon}
\newcommand{\Vcal}{\mathcal{V}}
\newcommand{\Ycal}{\mathcal{Y}}
\newcommand{\Greedy}{\mathcal{G}_\theta}
\newcommand{\Frep}{\widetilde{F}_{x,W}}
\newcommand{\ARR}{ARR}

\newcommand{\ARRar}{ARR\textsubscript{AR}}
\newcommand{\ARRpld}{ARR\textsubscript{PLD}}
\newcommand{\ARRdepth}{ARR\textsubscript{Depth}}
\newcommand{\FDR}{Self-Refine}
\newcommand{\VC}{SETS}
\newcommand{\SeqC}{Sequential Revision}
\newcommand{\SC}{Majority Voting}

\definecolor{cGreen}{RGB}{0,120,60}
\definecolor{cRed}{RGB}{180,30,30}
\newcommand{\cmark}{\textcolor{cGreen}{\ding{51}}}
\newcommand{\xmark}{\textcolor{cRed}{\ding{55}}}

\definecolor{bestcell}{RGB}{207,224,241}
\definecolor{secondcell}{RGB}{246,218,218}
\definecolor{oursrow}{RGB}{245,245,245}
\newcommand{\best}[1]{\cellcolor{bestcell}\textbf{#1}}
\newcommand{\second}[1]{\cellcolor{secondcell}#1}

\definecolor{boxText}{RGB}{20,20,20}
\definecolor{incorrectred}{RGB}{220,53,69}
\definecolor{correctgreen}{RGB}{40,167,69}
\definecolor{agentblue}{RGB}{0,123,255}
\definecolor{lightgray}{RGB}{248,249,250}
\definecolor{darkgray}{RGB}{108,117,125}
\definecolor{problemcolor}{RGB}{52,58,64}
\tcbset{arrstyle/.style={enhanced jigsaw, breakable, coltext=boxText,
  fonttitle=\bfseries\small, fontupper=\small, arc=2mm, outer arc=2mm, boxrule=1pt,
  left=5pt, right=5pt, top=4pt, bottom=4pt, before skip=6pt, after skip=8pt}}
\newcommand{\arrbox}[3]{\newtcolorbox{#1}[1]{arrstyle, colframe=#2, colback=#3, title={##1}}}
\arrbox{problembox}{problemcolor}{lightgray}
\arrbox{reasonbox}{agentblue}{blue!3!white}
\arrbox{critiquebox}{orange!80!black}{yellow!5!white}
\arrbox{correctionbox}{correctgreen}{green!5!white}
\arrbox{summarybox}{darkgray}{lightgray}
\arrbox{incorrectbox}{incorrectred}{red!5!white}
\definecolor{fieldcolor}{RGB}{120,40,160}
\definecolor{rolecolor}{RGB}{0,82,165}
\definecolor{goodcolor}{RGB}{0,115,45}
\definecolor{badcolor}{RGB}{185,25,40}
\definecolor{keyhl}{RGB}{255,232,130}
\newcommand{\ph}[1]{\textcolor{fieldcolor}{\textit{\{#1\}}}}
\newcommand{\role}[1]{\textcolor{rolecolor}{\textbf{#1:}}}
\newcommand{\ansfmt}[1]{\textcolor{goodcolor}{#1}}

\newcommand{\key}[1]{{\sethlcolor{keyhl}\hl{#1}}}
\newtcolorbox{observationbox}{arrstyle, colframe=orange!80!black, colback=yellow!5!white,
  title={Key Observations}}

\title{Adaptive Replay Refinement:\\Allocating Test-Time Compute Between Revision and Resampling}

\author{
  Ahsan Bilal\textsuperscript{1}\quad
  Muhammad Ahmed Mohsin\textsuperscript{2}\quad
  Muhammad Umer\textsuperscript{2} \\
  \textbf{Ayesha Mohsin\textsuperscript{3}\quad
  Sarmad Bilal\textsuperscript{3}\quad
  Dean F. Hougen\textsuperscript{1}} \\[4pt]
  \textsuperscript{1}University of Oklahoma\quad
  \textsuperscript{2}Stanford University\quad
  \textsuperscript{3}National University of Sciences and Technology
}

\begin{document}
\etocdepthtag.toc{mainmatter}
\maketitle
\begin{abstract}
Test-time scaling allocates additional inference compute to refining existing solutions or generating new ones. This raises a fundamental question: \emph{Should I keep improving this solution, or stop working on it and try a completely new solution?} Existing methods make this decision using fixed refinement schedules, self-verification judgments, or learned controllers, which can waste compute or rely on signals that may be incorrect or require additional training. We introduce \textbf{Adaptive Replay Refinement (ARR)}, a training-free method that detects when a stationary revision reaches a fixed point and reallocates the released compute to a new independent sample; replay decoding treats the current solution as a draft and verifies unchanged tokens in parallel, enabling efficient fixed-point detection. Across seven models from five families and seven benchmarks, ARR matches majority voting on three of five mathematical reasoning benchmarks at equal compute, while reducing critical-path serial passes by up to 2.4 times. On MATH500, ARR achieves 84.6\% accuracy with only 41\% of the serial passes required by majority voting. These results demonstrate that detecting refinement exhaustion enables adaptive reallocation from revision to independent sampling, improving inference efficiency while highlighting the limits of self-revision.\footnote{Code is available at \url{https://github.com/AhsanBilal7/adaptive_refine}.}
\end{abstract}

\begin{figure}
\centering
\includegraphics[width=\columnwidth]{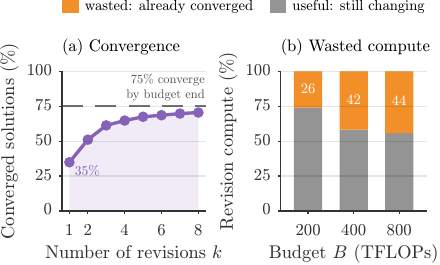}
\caption{Revision on MATH500 with Qwen2.5-Math-7B-Instruct. (a) Fraction of the four initial roots of \ARRdepth{} ($B{=}800$) at a fixed point: 35\% after one revision and 75\% by budget exhaustion. (b) Fraction of \FDR{} compute wasted on already converged solutions.}
\label{fig:motivation}
\end{figure}

\begin{figure*}
\centering
\includegraphics[width=\textwidth]{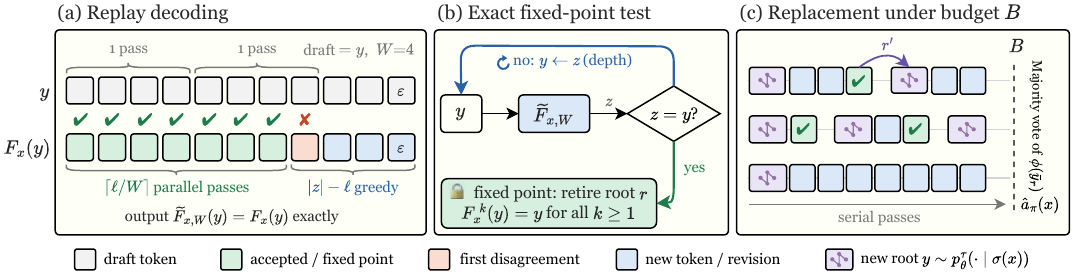}
\caption{Overview of ARR. (a)~Replay decoding accelerates revision. (b)~Exact reproduction identifies a fixed point and retires the root. (c)~Released compute is reallocated to new roots until the budget is exhausted.}
\label{fig:overview}
\end{figure*}

\section{Introduction} \label{sec:intro}
Large language models can improve reasoning performance by using more computation at inference time \citep{snell2024scalingllmtesttimecompute,wu2025inferencescalinglawsempirical,brown2024largelanguagemonkeysscaling}. Existing approaches primarily spend this compute in two ways: Parallel methods generate multiple independent solutions and aggregate their answers, typically through majority voting \citep{wang2023selfconsistencyimproveschainthought}, while sequential methods repeatedly revise a single solution, often through separate feedback and refinement stages \citep{madaan2023selfrefineiterativerefinementselffeedback,shinn2023reflexionlanguageagentsverbal}. Recent approaches combine these strategies by sampling several solutions, refining them, and aggregating the resulting answers \citep{chen2025setsleveragingselfverificationselfcorrection,snell2024scalingllmtesttimecompute}.

Combining sampling and refinement creates a fundamental allocation problem: When should additional compute try to improve an existing solution, and when should it generate a new one? Prior work typically relies on one of three mechanisms. Fixed refinement schedules apply a predetermined number of revision rounds, which can waste compute after further refinement stops changing a solution and often require validation data to choose the schedule \citep{madaan2023selfrefineiterativerefinementselffeedback}. Self-verification methods stop when the model judges its own solution to be correct \citep{chen2025setsleveragingselfverificationselfcorrection,pmlr-v306-samanta26a}, but such judgments can be unreliable without external feedback \citep{huang2024largelanguagemodelsselfcorrect,stechly2024selfverificationlimitationslargelanguage,kamoi2024llmsactuallycorrectmistakes}. Learned or score-based controllers instead use trained verifiers, predicted difficulty, or confidence estimates \citep{snell2024scalingllmtesttimecompute,wang-etal-2026-conma,fu2025deepthinkconfidence}, introducing additional training, calibration, or threshold selection. These approaches also incur unnecessary sequential computation because revision commonly regenerates most of a solution even when only a small portion changes. Figure~\ref{fig:motivation} shows that revision exhausts after few steps and that a fixed schedule wastes the rest of its compute.

We address this problem by detecting when a deterministic and stationary revision procedure can no longer change a solution. Once a revision exactly reproduces its input, further applications of the same procedure are redundant. We use this event as an exact signal that refinement of that solution has been exhausted. We introduce \textbf{Adaptive Replay Refinement (ARR)}, a training-free method that uses replay decoding to efficiently detect when refinement reaches a fixed point and reallocates the released compute to a new independent sample (Figure~\ref{fig:overview}). ARR therefore adapts refinement depth and sampling breadth directly during inference, without requiring a tuned refinement depth, correctness threshold, critic, or learned controller. While draft verification, fixed-point iteration, and training-free self-refinement have each been studied previously \citep{sun-etal-2021-instantaneous,pmlr-v202-leviathan23a,santilli2023acceleratingtransformerinferencetranslation,madaan2023selfrefineiterativerefinementselffeedback}, ARR uses these components to detect exhausted refinement and redirect compute to new attempts; a fixed point signals only that further revision is redundant, not that the solution is correct. Table~\ref{tab:positioning} in Appendix~\ref{app:related} compares this stop signal with those of the closest methods, and also discusses sequential scaling, efficient reasoning, and fixed points in sequence models.

\textbf{Contributions.}
First, we formulate iterative refinement as an \textbf{adaptive allocation problem} between improving existing solutions and generating new ones, and introduce a training-free rule that reallocates compute once refinement is exhausted. Second, we establish that \textbf{replay decoding preserves the output of ordinary greedy revision}, making exact reproduction a valid fixed-point stopping signal. Third, we introduce a \textbf{compute-matched evaluation protocol} that accounts for all processed computation and includes controlled variants that isolate the effect of each component. Fourth, across seven models from five model families and seven benchmarks, ARR matches majority voting on three of five mathematical reasoning benchmarks at equal compute while \textbf{shortening the serial critical path by up to $\mathbf{2.4}$ times}; our analysis also identifies two settings in which the underlying revision procedure itself is ineffective.

\section{Problem Setup}
\label{sec:setup}

Let $p_\theta$ be a frozen causal language model over vocabulary $\Vcal$, let $x$ be a question, and let $\Ycal\subset\Vcal^*$ denote the set of complete solutions $y$ that terminate with $\eos$ within the output token cap $\Tmax$.
Two solutions are equal only if they are identical as token sequences.
For a prompt $u$, the greedy map $\Greedy$ returns $z=\Greedy(u)$ with
\begin{equation}
z_t=\argmax_{v\in\Vcal}\, p_\theta\big(v \mid u, z_{<t}\big),
\label{eq:greedy}
\end{equation}
which stops at $\eos$ or after $\Tmax$ tokens.
A sample is $y\sim p_\theta^{\tau}(\cdot\mid\sigma(x))$, where $\sigma(x)$ is the solve prompt and $p_\theta^{\tau}$ is temperature-$\tau$ nucleus sampling.
The extractor $\phi:\Ycal\to\mathcal{A}$ returns the final answer of a solution, and $a^\star(x)$ is the correct answer.

\subsection{Allocating Test-Time Compute}
\label{sec:allocation}

A \emph{root} $r$ is one sample $y_r^{(0)}\sim p_\theta^{\tau}(\cdot\mid\sigma(x))$ together with its revisions $y_r^{(1)},y_r^{(2)},\dots$, and $\bar y_r$ is its last complete solution.
A method receives a per-question budget $B$ in FLOPs.
At each step, its allocation policy $\pi$ revises an active root or samples a new root.
The policy $\pi$ allocates the budget $B$ between revising existing roots and sampling new ones, stopping when no further action fits. The final prediction is the majority vote over complete roots, $\hat a_\pi(x)=\argmax_{a\in\mathcal{A}}\sum_r \mathds{1}[\phi(\bar y_r)=a].$ Thus, the objective is
\begin{equation}
\max_{\pi}\ \Pr\big[\hat{a}_\pi(x)=a^\star(x)\big]
\quad \text{s.t.} \quad
\mathcal{C}_\pi(x)\le B .
\label{eq:objective}
\end{equation}
Since $a^\star(x)$ is unknown at inference time, $\pi$ must rely on observable signals; ARR uses the fixed-point condition $F_x(y)=y$, which holds when the deterministic revision operator exactly reproduces its input, implying that further applications of the same operator leave the solution unchanged.

\subsection{Stationary Revision Operator} \label{sec:operator}
Let $\rho(x,y)$ be a fixed revision prompt containing the question $x$ and solution $y$, and define the stationary revision operator as $F_x(y)=\Greedy(\rho(x,y))$. The prompt asks the model to correct any errors and their downstream consequences while preserving correct content, and to reproduce $y$ verbatim when no correction is needed (Appendix~\ref{app:prompts}). ARR relies on three properties of $F_x$: it uses a \emph{single call} that combines critique and correction into one revised solution \citep{chen2025setsleveragingselfverificationselfcorrection}; it is \emph{stationary} because $\rho$ depends only on $(x,y)$ and contains no round index, revision history, or other roots, unlike prompts that change across rounds \citep{madaan2023selfrefineiterativerefinementselffeedback}; and it is \emph{deterministic} under greedy decoding, making $F_x(y)=y$ well defined and confining diversity to sampled roots, consistent with evidence that LLM representations encode future response attributes before generation \citep{pmlr-v267-dong25p}.

\section{Adaptive Replay Refinement} \label{sec:method}
ARR is a training-free policy for Equation~(\ref{eq:objective}).
It revises each root with $F_x$ to a fixed point and then gives the slot of the root to a new sample as shown in Figure~\ref{fig:overview}.
Replay decoding computes $F_x(y)$ with $y$ as the draft, so the fixed-point test costs few serial passes (Sections~\ref{sec:replay} and~\ref{sec:exact}), and this test drives the allocation (Section~\ref{sec:replace}).
ARR has three settings: the width $N$, i.e., the number of concurrently active roots, the budget $B$, and the block size $W$.
Algorithms~\ref{alg:replay} and~\ref{alg:arr} in Appendix~\ref{app:algorithms} give the pseudocode.

\subsection{Replay Decoding}
\label{sec:replay}

A revision usually reproduces a long prefix of its input, but greedy decoding spends one serial forward pass on each reproduced token.
Replay decoding uses the input $y$ as a draft and verifies up to $W$ draft positions in one pass.

\textbf{Block verification.} Let $u=\rho(x,y)$, and assume that the committed output agrees with the draft, $z_{1:t}=y_{1:t}$.
Replay takes the next $m=\min(W-1,\,|y|-t)$ draft tokens and runs one causal pass on $(z_t,y_{t+1},\dots,y_{t+m})$, which gives, for $j=1,\dots,m{+}1$, the greedy predictions
\begin{equation}
\hat z_{t+j} = \argmax_{v\in\Vcal}\, p_\theta\big(v \mid u, y_{1:t+j-1}\big).
\label{eq:block}
\end{equation}
Replay accepts the longest prefix of the block on which the predictions agree with the draft, of length $\lambda=\max\{i\le m: \hat z_{t+j}=y_{t+j}\ \forall j\le i\}$.
It appends $y_{t+1:t+\lambda}$ and the model token $\hat z_{t+\lambda+1}$ to $z$, and it discards the cache of the rejected positions.
If $\hat z_{t+\lambda+1}=y_{t+\lambda+1}$, the output is still on the draft, and replay verifies the next block.
Otherwise, the revision left the draft, and replay decodes the rest with Equation~(\ref{eq:greedy}).

\textbf{Design choices.} Replay verifies only a contiguous draft prefix, since tokens beyond the first mismatch are conditioned on an invalid prefix. Unlike aggressive and prompt lookup decoding, which may realign after mismatches \citep{sun-etal-2021-instantaneous,saxena2023pld} (Section~\ref{sec:exp-setup}), replay switches immediately to standard greedy decoding, preserving exact equivalence to $F_x$ while incurring at most one failed block per revision. Thus, replay verifies the existing solution against a new revision, and when the entire draft is accepted without changes, ARR detects a fixed point and reallocates compute to a new sample.

\subsection{Exact Fixed-Point Detection}
\label{sec:exact}
Let $\Frep(y)$ be the output of replay decoding. Replay changes only how the revision is computed, not its output: under deterministic argmax decoding with fixed tie-breaking, $\Frep(y)=F_x(y)$ for every $y\in\Ycal$ and $W\ge1$, since each verified token matches the greedy continuation in Equation~(\ref{eq:greedy}). Because $\rho$ depends only on $(x,y)$ and greedy decoding is deterministic, $\Frep(y)=y$ implies $F_x^{\,k}(y)=y$ for all $k\ge1$. Thus, accepting the entire draft, including $\eos$, exactly identifies an exhausted revision, while any disagreement implies $F_x(y)\neq y$. Appendix~\ref{app:impl} provides the implementation details and backend validation across four architectures.

\textbf{Serial cost.} Let $z=F_x(y)$, and let $\ell$ be the length of the longest common prefix of $z$ and $y$.
Greedy decoding uses $1+|z|$ serial passes.
Replay uses about $1+\lceil \ell/W \rceil+(|z|-\ell)$, because it verifies the reproduced prefix in blocks and decodes only the changed suffix token by token.
A fixed point with $|z|=512$ and $W=32$ thus takes 17 passes instead of 513.
The processed positions, and thus the FLOPs, stay almost the same, because a revision discards at most $W{-}1$ positions.
The saving is in serial depth, which sets the latency.

\subsection{Fixed-Point Replacement}
\label{sec:replace}

ARR starts with $N$ roots, one for each slot.
Let $z=\Frep(y_r^{(k)})$ be the revision of an active root $r$ at depth $k$.
ARR applies
\begin{equation}
\begin{cases}
y_r^{(k+1)} = z, & z\in\Ycal\setminus\{y_r^{(k)}\},\\[2pt]
\bar y_r = y_r^{(k)},\ \ \mathcal{R} \leftarrow \mathcal{R}\cup\{r'\}, & z = y_r^{(k)},\\[2pt]
\bar y_r = y_r^{(k)}, & z\notin\Ycal,
\end{cases}
\label{eq:rule}
\end{equation}
where $\mathcal{R}$ is the set of roots and $r'$ is a new root, sampled from $\sigma(x)$, that takes the slot of $r$. In the first case, the solution still changes, and ARR revises it again.
In the second case, $y_r^{(k)}$ is a fixed point (Section~\ref{sec:exact}): its answer becomes the final vote of the root, and the slot samples a new root if a complete sample still fits in $B$.
In the third case, $\Tmax$ or the budget truncated the revision: the root keeps its last complete solution, and ARR does not refill the slot.

Equation~(\ref{eq:rule}) is the full allocation policy, with no depth limit, threshold, score, or learned parameter.
A new root does not observe the retained answers, making it an independent attempt; ARR returns the majority answer $\hat a_\pi(x)$ of Section~\ref{sec:allocation}, with Appendix~\ref{app:example} illustrating replay and replacement through a worked example.

\subsection{Compute Accounting}
\label{sec:budget}

We use a single FLOP accounting for all methods. The total cost of a method on question $x$ is the sum of the FLOPs of every processed token position, $ \mathcal{C}_\pi(x)=\sum_{i\in\mathcal{P}_\pi(x)}\kappa(c_i),$ where $\kappa(c_i)$ is the FLOP cost of processing position $i$ when it attends to $c_i$ cached keys (Appendix~\ref{app:ledger}). This accounting includes all computation: prompt prefills, verified draft tokens, discarded verification work, decoded tokens, and truncated calls. Thus, no replay or revision computation is treated as free. We separately measure latency by the number of serial forward passes $S$ on the critical path of each question.

\section{Experiments} \label{sec:experiments}
We describe the setup in Section~\ref{sec:exp-setup}, present the main results in Section~\ref{sec:results}, analyze when revision should receive compute in Section~\ref{sec:dynamics}, analyze individual components in Section~\ref{sec:ablation}, evaluate generality across models in Section~\ref{sec:models}, and examine failure modes in Section~\ref{sec:failure}.

\begin{figure*}
\centering
\includegraphics[width=0.9\textwidth]{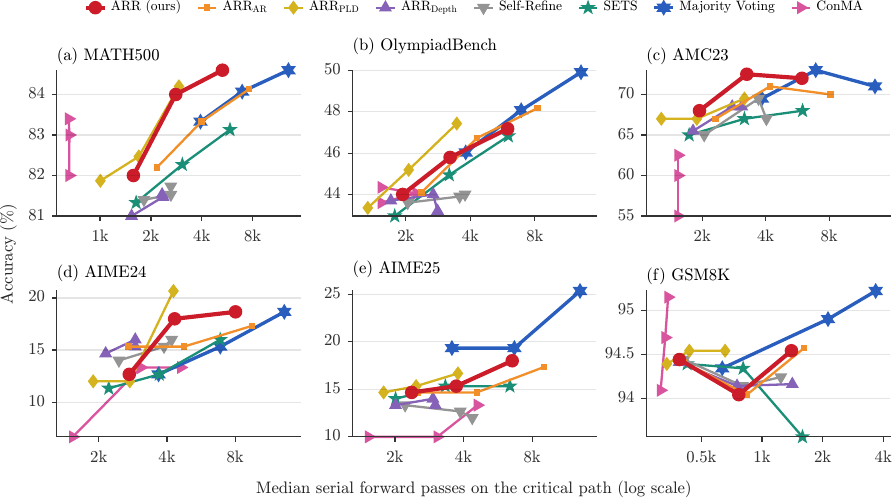}
\caption{Accuracy against median critical-path serial passes (Qwen2.5-Math-7B-Instruct); each point is one budget.}
\label{fig:main}
\end{figure*}

\subsection{Setup} \label{sec:exp-setup}
\textbf{Benchmarks.} We evaluate on five mathematical reasoning benchmarks with exact-answer scoring: MATH500 \citep{hendrycks2021measuringmathematicalproblemsolving} (500 problems), AIME (30 problems per edition), AMC (40 problems), and the 674 English text-only competition mathematics problems from OlympiadBench \citep{he2024olympiadbenchchallengingbenchmarkpromoting}. We include GSM8K \citep{cobbe2021trainingverifierssolvemath} (1{,}319 problems) as a saturated control and evaluate generalization beyond mathematics on LiveBench reasoning tasks \citep{white2025livebenchchallengingcontaminationlimitedllm}, comprising web of lies, zebra puzzles, and spatial reasoning (50 problems each). Appendix~\ref{app:decoding} details answer extraction and scoring.

\textbf{Models.} The primary model is Qwen2.5-Math-7B-Instruct \citep{yang2024qwen25mathtechnicalreportmathematical}. For breadth, we evaluate Qwen3-8B \citep{yang2025qwen3technicalreport}, Ministral-8B-Instruct-2410 \citep{mistral2024ministral}, Gemma-4-E4B-it \citep{gemma2026gemma4}, and Qwen3.5-9B and 2B \citep{qwen2026qwen35}. These models span multiple families, scales, and decoding-state architectures, including full, sliding-window, and hybrid linear attention, testing whether replay-based fixed-point detection generalizes across inference backends rather than a specific cache design. For the 7-8B models, we use budgets of $B\in\{200,400,800\}$ TFLOPs; for models with longer generations, budgets are scaled to preserve comparable sampling capacity across budget levels (Appendix~\ref{app:budgets} converts each budget to a number of samples), with FLOPs computed using the common FLOP accounting in Appendix~\ref{app:ledger}. We additionally evaluate Qwen2.5-7B-Instruct \citep{qwen2025qwen25technicalreport} on LiveBench.

\textbf{Compared methods.}
We select baselines spanning independent sampling, fixed-depth and stochastic revision, explicit self-verification, and confidence-based allocation, alongside controlled ARR variants that isolate decoding and root replacement. All methods use the same FLOP accounting, $\Tmax$, and answer extractor $\phi$; root-based methods additionally share $N{=}4$ initial roots and majority aggregation, while ConMA retains its group-based selection procedure (Table~\ref{tab:methods}, Appendix~\ref{app:methods}). \textbf{\SC{}} \citep{wang2023selfconsistencyimproveschainthought} spends the entire budget on independent samples from $\sigma(x)$, providing a pure-sampling baseline for evaluating the benefit of revision. \textbf{\ARRar{}} and \textbf{\ARRpld{}} retain ARR's allocation rule but compute $F_x$ with autoregressive greedy decoding and prompt lookup decoding \citep{saxena2023pld}, respectively, isolating the effect of replay from the underlying allocation policy; their single revisions are equivalent to ARR's under Section~\ref{sec:exact}, while full runs can differ because of decoding cost and scheduling (Appendix~\ref{app:methods}). \textbf{\ARRdepth{}} removes root replacement from Equation~(\ref{eq:rule}) to isolate the effect of reallocating exhausted depth to breadth, while \textbf{\FDR{}} \citep{madaan2023selfrefineiterativerefinementselffeedback} applies $F_x$ for a fixed $D{=}4$ revisions, providing an iteration-limited baseline for assessing redundant refinement. \textbf{\VC{}} \citep{chen2025setsleveragingselfverificationselfcorrection} replaces $F_x$ with a two-stage critique-and-correction update, $F_x^{\mathrm{vc}}$, while keeping ARR's replacement rule, isolating the effect of explicit self-verification on revision quality. \SeqC{} \citep{snell2024scalingllmtesttimecompute} repeatedly samples a non-stationary continuation from $\rho_{\mathrm{cont}}(x,y_r^{(k)},k)$ until the budget is exhausted, contrasting stochastic, depth-dependent revision with ARR's deterministic stationary operator. \textbf{ConMA} \citep{wang-etal-2026-conma} is a confidence-based controller: it samples groups of trajectories, keeps the kernel of answers with the highest token-level confidence, and re-samples or asks a multiple-choice question until one answer remains, providing an adaptive alternative based on confidence rather than exact fixed-point detection (Appendix~\ref{app:methods}). We fix $N{=}4$ for root-based methods and $W{=}32$ for ARR's replay decoder across models without tuning; replay exactness (Section~\ref{sec:exact}) ensures that $W$ affects decoding speed but not revision output.

\textbf{Metrics.} We report accuracy, mean TFLOPs spent, and median serial passes $S$ per question, with 95\% bootstrap confidence intervals computed over question-paired differences relative to ARR. All experiments are run with multiple random seeds (three to five per setting), and all reported results are averaged over seeds (Table~\ref{tab:stop}: paired seeds only). Using matched initial roots, we define the \emph{net repair} of $F_x$ as the difference between incorrect-to-correct and correct-to-incorrect answer changes, normalized by the number of complete revisions. Appendix~\ref{app:impl} details the mechanism statistics, decoder, and hardware.

\subsection{Main Results} \label{sec:results}
\textbf{ARR gives more accuracy for each serial pass.} Figure~\ref{fig:main} shows accuracy against the median number of serial passes on the critical path. This number sets the latency of a question.
On MATH500, AMC23, and AIME24, the curve of ARR is above the curves of \ARRar{}, \VC{}, and \SC{} for most of the range.
For example, ARR gets the best accuracy of \SC{} on MATH500 (84.6\%) with only 41\% of the serial passes of \SC{}.
On OlympiadBench, the curves of ARR and \SC{} are almost the same, and on AIME25 and GSM8K, \SC{} is better.

\textbf{At equal FLOPs, ARR matches majority voting on three of five math benchmarks.} At the largest budget, ARR and \SC{} have the same accuracy within noise on MATH500, AMC23, and AIME24 (Figure~\ref{fig:budget}, Appendix~\ref{app:results}).
On OlympiadBench and AIME25, \SC{} is better by up to 7.3 points.
The cause is the number of votes. \SC{} spends all of the budget on independent samples, but ARR also spends compute on revisions and thus collects approximately six times fewer votes (Table~\ref{tab:cost}).
More votes help most when revision repairs few answers (Section~\ref{sec:dynamics}).
All other methods that revise get lower accuracy, and \ARRdepth{} and \FDR{} also spend fewer FLOPs (Figure~\ref{fig:compute}).

\textbf{ARR and majority voting give different trade-offs.}\label{sec:latency} Each slot of \SC{} samples its votes in sequence, so its critical path is long.
At equal FLOPs, ARR needs only 41-61\% of the serial passes of \SC{}.
On one A100 GPU and a MATH500 subset, ARR answers a question in a median of 78.5\,s, and \SC{} answers it in 176.5\,s with the same accuracy (Table~\ref{tab:timing}, Appendix~\ref{app:results}).
At equal latency, the result changes: with a limit of 8{,}000 serial passes on MATH500, the two methods get the same accuracy, but \SC{} uses less than half of the FLOPs (Figure~\ref{fig:abl-passes}).
Thus, ARR is more efficient in latency at a fixed FLOP budget, and \SC{} is more efficient in FLOPs at a fixed latency.

\textbf{Replay makes revision faster without a change in output.} Replay gives the same revisions as \ARRar{} and decreases the critical path by up to 30\%.
This decrease becomes larger with the budget, because deeper roots reproduce longer prefixes (Figure~\ref{fig:replay}).
Because replay verifies reused text in parallel, ARR also writes fewer tokens than \SC{}: on MATH500, it writes approximately 40\% of the output tokens of \SC{} for the same accuracy.

\textbf{A confidence controller saves compute but loses accuracy.} ConMA stops generating samples once only one answer remains under consideration, reducing inference cost through early stopping. Across completed settings, it uses approximately half of the compute budget and requires half as many serial passes as ARR (Figure~\ref{fig:compute}). However, this efficiency comes at the cost of accuracy: at $B{=}800$, ConMA performs 1.2-9.5 percentage points worse than ARR (Figure~\ref{fig:budget}). This suggests that confidence-based stopping can save compute but may terminate sampling too early, potentially missing better answers.

\subsection{When Revision Should Receive Compute} \label{sec:dynamics}

\begin{figure*}[t]
\centering
\includegraphics[width=0.95\textwidth]{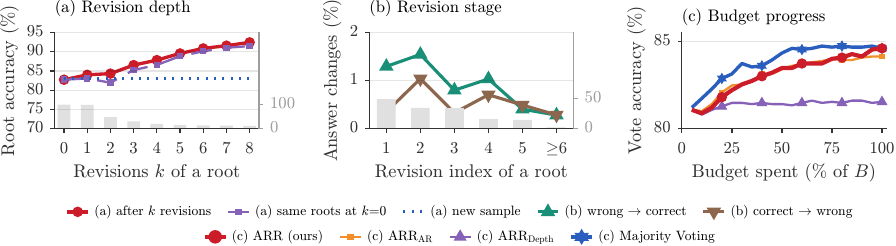}
\caption{Revision in ARR on MATH500 ($B{=}800$). (a) Accuracy of roots by revision depth; bars: roots that reach depth $k$. (b) Answer changes per revision; bars: fixed-point rate. (c) Accuracy of the running vote.}
\label{fig:dynamics}
\end{figure*}

\begin{figure*}[t]
\centering
\includegraphics[width=0.95\textwidth]{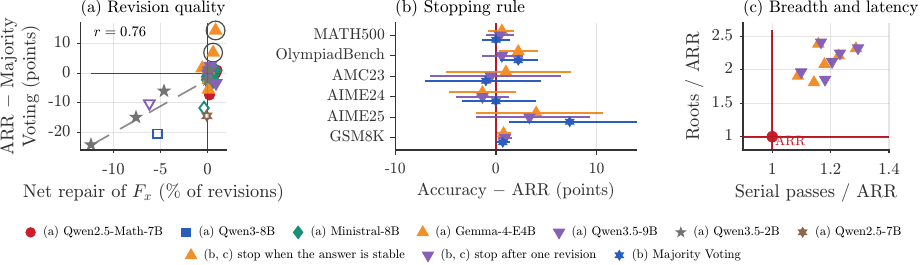}
\caption{(a) Net repair of $F_x$ vs. ARR $-$ \SC{}; open = LiveBench, circled = repaired answer format. (b) Accuracy of the other stopping rules and \SC{} relative to ARR, with 95\% CIs. (c) Their roots and serial passes relative to ARR.}
\label{fig:regimes}
\end{figure*}

\textbf{Revision provides early gains but quickly saturates.}
Revision improves root accuracy by 0.5-2.4 points, but 49\% of the first revisions of ARR (over all roots, including new ones) reach a fixed point, and answer changes become rare after four revisions (Figure~\ref{fig:dynamics}).
By reallocating compute from converged roots to new samples, ARR reaches 84.6\% accuracy, matching \SC{} and outperforming \ARRdepth{} (81.5\%; Figure~\ref{fig:dynamics}).

\textbf{The fixed point is an exact signal, not necessarily the most accurate stop.}
ARR's condition $F_x(y)=y$ guarantees that further deterministic revisions leave the solution unchanged, without requiring answer extraction, thresholds, or correctness labels. We compare two simpler rules with the same root replacement: (i) answer stability, which stops when $\phi(F_x(y))=\phi(y)$, and (ii) one-revision stopping, which retires a root after a single revision. At $B{=}800$, answer stability outperforms ARR by 4.0 points on AIME25 (95\% CI $[-2.0,+10.7]$) and 2.2 on OlympiadBench ($[+0.3,+4.2]$), while both stopping rules remain within 1 point on MATH500 and GSM8K (Figure~\ref{fig:regimes}(b), Table~\ref{tab:stop}). These rules retire roots before their solutions converge, generating 1.8-2.4 times more roots but requiring 1.09-1.29 times more serial passes (Figure~\ref{fig:regimes}(c)). Thus, fixed-point stopping gives the shortest critical path among the tested rules, although earlier stopping can sometimes improve accuracy.

\textbf{The net repair of $F_x$ decides between revision and sampling.} Over different model-benchmark settings, the net repair of $F_x$ predicts the accuracy difference between ARR and \SC{} (Pearson $r{=}0.76$, Figure~\ref{fig:regimes}(a)). With a net repair above 0.25\%, ARR is 1.3 points better than \SC{} on average; with a negative net repair, \SC{} is 8.7 points better, most of all on LiveBench and with Qwen3.5-2B (Section~\ref{sec:failure}). Revision should thus receive compute only when it repairs more answers than it breaks, and this condition explains both the successes and the failures of ARR. In two Gemma-4 settings, ARR is 7.0-14.5 points better for a reason other than reasoning: 37-41\% of the samples end with a boxed answer, against 98-100\% after revision.

\subsection{Component Analysis} \label{sec:ablation}
\textbf{The exchange of depth for breadth is the source of the gain.} \ARRdepth{} keeps the fixed-point test but does not replace retired roots.
Thus, it stops when its four roots get to fixed points, and it uses less than half of the budget on MATH500.
ARR gives this released compute to new roots and is 3.1 points more accurate on MATH500 and 4.0 points more accurate on OlympiadBench.
Figure~\ref{fig:depthacc} shows the two sources of gain separately. Revision of the initial roots adds at most 2.5 points.
The new roots add 3.0-6.0 points, and this gain increases with the budget. The same order holds for all models.

\textbf{A greedy and stationary revision operator works best.} We compare two revision alternatives (Table~\ref{tab:methods}) across FLOP budgets (Figure~\ref{fig:budget}). \VC{} adds an explicit critique step, reducing MATH500 accuracy by 1.5 points while remaining within noise on other benchmarks. Despite reaching more fixed points (69\% vs.\ 23\%), it repairs fewer answers (Table~\ref{tab:cost}). \SeqC{} replaces greedy revision with sampled continuation, causing MATH500 accuracy to drop from 79.3\% to 59.0\% as the budget increases, 25.6 points below ARR at $B{=}800$ (Figure~\ref{fig:budget}(a)). It also breaks more correct answers than it repairs (Table~\ref{tab:cost}).

\textbf{The allocation rule does not depend on replay.} \ARRpld{} uses the same rule with prompt lookup decoding. It gets similar accuracy with approximately 55\% of the serial passes of ARR. The cause is the first change in a revision. On average, a revision reproduces only 26-35\% of its input before this change. After this point, replay decodes one token at a time, but prompt lookup can align to the draft again. Thus, a decoder with realignment makes ARR faster, and the fixed-point signal stays the same. In wall-clock time, \ARRpld{} needs 54\% of the latency of ARR (Table~\ref{tab:timing}, Appendix~\ref{app:results}).

\textbf{ARR is robust to its settings.} A larger width helps: with $N{=}8$, ARR gains 1.9 points over $N{=}4$ and needs fewer serial passes. The block size $W$ changes the serial passes by at most 5\%, and two paraphrases of $\rho$ change the accuracy by at most 1.2 points. Appendix~\ref{app:results} gives all ablations and the detailed values of this section.

\subsection{Generality across Models}
\label{sec:models}

\textbf{The results transfer to four more model families.} Figure~\ref{fig:models} shows the five other models at their largest budget, over different settings. Three trends of the primary model also hold for these models.
First, the exchange of depth for breadth helps: ARR is better than \ARRdepth{} and \FDR{} in most of the settings, by approximately 6 points on average.
Second, the decoder does not change accuracy: ARR, \ARRar{}, and \ARRpld{} are within 0.4 points.
Third, ARR and \SC{} are close, and \SC{} is 1.5 points better on average.
This difference decreases to 0.7 points when we remove Qwen3.5-2B, where revision breaks answers (Section~\ref{sec:failure}), and two Gemma-4 settings, where revision only repairs the answer format.

\subsection{Failure Modes}
\label{sec:failure}
A fixed point shows that revision is exhausted, not that the solution is correct. Thus, ARR is weak when revision breaks more answers than it repairs. On LiveBench, \SC{} is better for all five models, by up to 20.4 points (Figures~\ref{fig:models}(f) and~\ref{fig:regimes}(a)). With Qwen3.5-2B, revisions are very short and often give only a final answer, and ARR is 15-24 points below \SC{} on MATH500 and AMC23 (Figure~\ref{fig:models}(a,c)). The accuracy of ARR is therefore limited by the quality of its revision operator.

\section{Related Work} \label{sec:related}
We review parallel test-time scaling and sample aggregation in Section~\ref{sec:parallel-scaling}, self-refinement and self-correction in Section~\ref{sec:self-refinement}, and draft-and-verify decoding in Section~\ref{sec:draft-verify}.

\subsection{Parallel Test-Time Scaling and Sample Aggregation} \label{sec:parallel-scaling}
Trained verifiers can rank candidate solutions \citep{cobbe2021trainingverifierssolvemath,lightman2023letsverifystepstep}, whereas self-consistency samples multiple chain-of-thought solutions and returns the most frequent answer \citep{wang2023selfconsistencyimproveschainthought}.
Adaptive-Consistency \citep{aggarwal2023letssamplestepstep} and ESC \citep{li2024escapeskyhighcostearlystopping} terminate sampling once answer agreement is sufficiently strong.
\citet{snell2024scalingllmtesttimecompute} allocate inference compute between sampling and revision using a learned verifier and predicted problem difficulty, and solution coverage generally increases as more samples are generated \citep{brown2024largelanguagemonkeysscaling,wu2025inferencescalinglawsempirical}.
DeepConf \citep{fu2025deepthinkconfidence} and ConMA \citep{wang-etal-2026-conma} instead use token-level confidence signals to filter or expand candidate groups.

\subsection{Self-Refinement and Self-Correction}\label{sec:self-refinement}
\citet{welleck2022generatingsequenceslearningselfcorrect} train a separate corrector that iteratively revises the outputs of a generator.
Reflexion \citep{shinn2023reflexionlanguageagentsverbal} retains verbal feedback across trials, while Self-Refine \citep{madaan2023selfrefineiterativerefinementselffeedback} alternates self-generated feedback with iterative revision.
Self-Debug \citep{chen2023teachinglargelanguagemodels} and CRITIC \citep{gou2024criticlargelanguagemodels} incorporate execution feedback or external tools.
Without such feedback, intrinsic self-correction often fails to improve reasoning and planning performance \citep{huang2024largelanguagemodelsselfcorrect,stechly2024selfverificationlimitationslargelanguage,kamoi2024llmsactuallycorrectmistakes}.
Other approaches train multi-turn self-correction directly through supervised or reinforcement learning \citep{qu2024recursiveintrospectionteachinglanguage,kumar2024traininglanguagemodelsselfcorrect}.
SETS \citep{chen2025setsleveragingselfverificationselfcorrection} repeatedly verifies and corrects sampled solutions until receiving a ``correct'' verdict before aggregation, whereas Thought-ICS \citep{pmlr-v306-samanta26a} identifies the first erroneous reasoning step and resamples from the preceding prefix.

\subsection{Draft-and-Verify Decoding}\label{sec:draft-verify}
Aggressive decoding \citep{sun-etal-2021-instantaneous} verifies spans copied from the input in parallel.
Speculative decoding accelerates autoregressive generation by drafting multiple tokens and verifying them with the target model in parallel \citep{pmlr-v202-leviathan23a,chen2023acceleratinglargelanguagemodel}.
LLMA \citep{yang2023inferencereferencelosslessacceleration} reuses spans from a reference sequence, while prompt lookup decoding \citep{saxena2023pld} and PLD+ \citep{somasundaram-etal-2025-pld} retrieve candidate continuations directly from the prompt.

\section{Conclusion}
\label{sec:conclusion}
ARR reframes test-time scaling as an online allocation problem between improving an existing trajectory and exploring a new one. Our results suggest that allocation does not require predicting correctness directly; detecting when refinement stops being useful can be enough to redirect compute to new attempts. This perspective opens several directions, including stronger or task-adaptive revision operators, decoders that realign after edits, integration with verifiers and reasoning models, and richer signals for deciding when to revise, resample, or terminate. More generally, progress-aware allocation could enable inference systems to allocate compute based on the marginal utility of continued reasoning, prioritizing useful reasoning over fixed schedules.

\section*{Limitations}
ARR requires access to model logits and the decoding loop, limiting its use with text-only APIs or hidden-reasoning systems. A fixed point indicates only that deterministic refinement is exhausted, not that the solution is correct, so ARR relies on independent replacement roots and aggregation. Finally, ARR prioritizes lower serial latency over matched-FLOP accuracy; its weaker regimes on LiveBench and Qwen3.5-2B motivate evaluating stronger revision operators and larger reasoning models.

\subsection*{AI Use Statement}

\texttt{ChatGPT} by \texttt{OpenAI} and \texttt{Claude} by \texttt{Anthropic} were used for minor language editing, proofreading, and resolving \LaTeX{} formatting issues. \texttt{Codex} by \texttt{OpenAI} assisted with routine code implementation, refactoring, plotting, and table generation for Adaptive Replay Refinement (ARR). All AI-assisted content and code were reviewed and verified by the authors, who take full responsibility for the methodology, experimental results, and manuscript.

\subsection*{Ethics Statement}

This work investigates test-time compute allocation between iterative self-refinement and independent solution sampling using publicly available language models and established reasoning benchmarks. The study involves no human subjects, collection of personally identifiable information, or new dataset creation.

\bibliography{custom}

\clearpage
\appendix

\renewcommand{\topfraction}{0.75}
\renewcommand{\dbltopfraction}{0.75}
\renewcommand{\bottomfraction}{0.8}
\renewcommand{\textfraction}{0.2}
\renewcommand{\floatpagefraction}{0.8}
\renewcommand{\dblfloatpagefraction}{0.8}
\setcounter{topnumber}{3}
\setcounter{dbltopnumber}{2}
\setcounter{bottomnumber}{2}
\setcounter{totalnumber}{5}

\etocdepthtag.toc{appendix}
\etocsettagdepth{mainmatter}{none}
\etocsettagdepth{appendix}{subsection}
\etocsettocstyle{\section*{Appendix Contents}}{}
\tableofcontents

\section{Pseudocode}
\label{app:algorithms}

Algorithm~\ref{alg:replay} computes one revision with replay decoding, and Algorithm~\ref{alg:arr} runs the allocation rule of Equation~(\ref{eq:rule}) under the budget $B$.
Jobs of all roots share one first-in-first-out queue.
An initial sample that is not complete is charged but never votes, and its slot samples again (\textsc{Spawn}).

\begin{algorithm}[tp]
\caption{\textsc{Replay}$(x, y, W)$ (Section~\ref{sec:replay})}
\label{alg:replay}
\footnotesize
\begin{algorithmic}[1]
\Require question $x$, solution $y\in\Ycal$, block size $W$
\State $u \gets \rho(x,y)$;\ prefill $u$;\ $z \gets \big(\argmax_v p_\theta(v \mid u)\big)$
\State $\mathit{on} \gets [\,z_1 = y_1\,]$ \Comment{$z$ is a prefix of $y$}
\While{$z_{|z|} \neq \eos$ \textbf{and} $|z| < \Tmax$}
  \State $t \gets |z|$
  \If{$\mathit{on}$}
    \State $m \gets \min(W{-}1, |y|{-}t)$
    \State pass $(z_t, y_{t+1:t+m}) \to \hat z_{t+1:t+m+1}$ \Comment{Eq.~(\ref{eq:block})}
    \State $\lambda \gets$ accepted length \Comment{Section~\ref{sec:replay}}
    \State $z \gets z \oplus y_{t+1:t+\lambda}$
    \State discard cache of $y_{t+\lambda+1:t+m}$
    \If{$z_{|z|} \neq \eos$}
      \State $z \gets z \oplus \hat z_{t+\lambda+1}$
      \State $\mathit{on} \gets [\,\hat z_{t+\lambda+1} = y_{t+\lambda+1}\,]$
    \EndIf
  \Else
    \State $z \gets z \oplus \argmax_v p_\theta(v \mid u, z)$ \Comment{Eq.~(\ref{eq:greedy})}
  \EndIf
\EndWhile
\State \Return $z$
\end{algorithmic}
\end{algorithm}

\begin{algorithm}[tp]
\caption{\textsc{ARR}$(x, N, B, W)$ (Section~\ref{sec:replace})}
\label{alg:arr}
\footnotesize
\begin{algorithmic}[1]
\Require question $x$, width $N$, budget $B$, block size $W$
\State $\mathcal{R} \gets \{1,\dots,N\}$
\State FIFO queue $Q \gets (\textsc{Sample}(r))_{r \in \mathcal{R}}$
\While{$Q \neq \emptyset$ \textbf{and} the next job fits in $B$}
  \State $j \gets Q.\mathrm{pop}()$
  \If{$j = \textsc{Sample}(r)$}
    \State $y_r \sim p_\theta^{\tau}(\cdot \mid \sigma(x))$
    \If{$y_r \in \Ycal$}
      \State push \textsc{Revise}$(r)$
    \Else
      \State \textsc{Spawn}() \Comment{invalid sample}
    \EndIf
  \Else \Comment{$j = \textsc{Revise}(r)$}
    \State $z \gets \textsc{Replay}(x, y_r, W)$ \Comment{Algorithm~\ref{alg:replay}}
    \If{$z \notin \Ycal$}
      \State $\bar y_r \gets y_r$ \Comment{truncated: freeze}
    \ElsIf{$z = y_r$}
      \State $\bar y_r \gets y_r$;\ \textsc{Spawn}() \Comment{fixed point}
    \Else
      \State $y_r \gets z$;\ push \textsc{Revise}$(r)$
    \EndIf
  \EndIf
\EndWhile
\State \Return majority answer $\hat a_\pi(x)$ \Comment{Section~\ref{sec:allocation}}
\Statex \textsc{Spawn}(): if a complete sample fits in $B$, add a root $r'$ to $\mathcal{R}$ and push \textsc{Sample}$(r')$.
\end{algorithmic}
\end{algorithm}

\section{Decoding, Formatting, and Answer Extraction}
\label{app:decoding}

\textbf{Termination.} The token $\eos$ is the end-of-sequence token of the chat template of each model.
Because $\eos$ is part of the sequence, the test $z=y$ also compares the termination decision: a revision that stops earlier or later than $y$ is not a fixed point.

\textbf{Prompting and precision.} All models run in bfloat16 with their own chat templates and with thinking modes disabled.

\textbf{Sampling.} Sampled calls use temperature $\tau{=}0.7$ and nucleus threshold $0.95$.
Each request draws its random numbers from its own seeded generator, so its samples do not depend on the other requests in a batch.
Greedy calls use argmax with the tie-breaking of the backend.

\textbf{Answer extraction and voting.} For math, $\phi$ returns the content of the last \verb|\boxed{}|.
For LiveBench, the answer line follows the format of each task (Appendix~\ref{app:prompts}), and the official LiveBench scripts score the answers.
Intermediate revisions do not vote.
A tie in the majority vote goes to the root with the lowest index, and a question without a vote counts as incorrect.

\section{Compute Accounting and Budgets}
\label{app:ledger}

\begin{figure*}[t]
\centering
\includegraphics[width=0.9\textwidth]{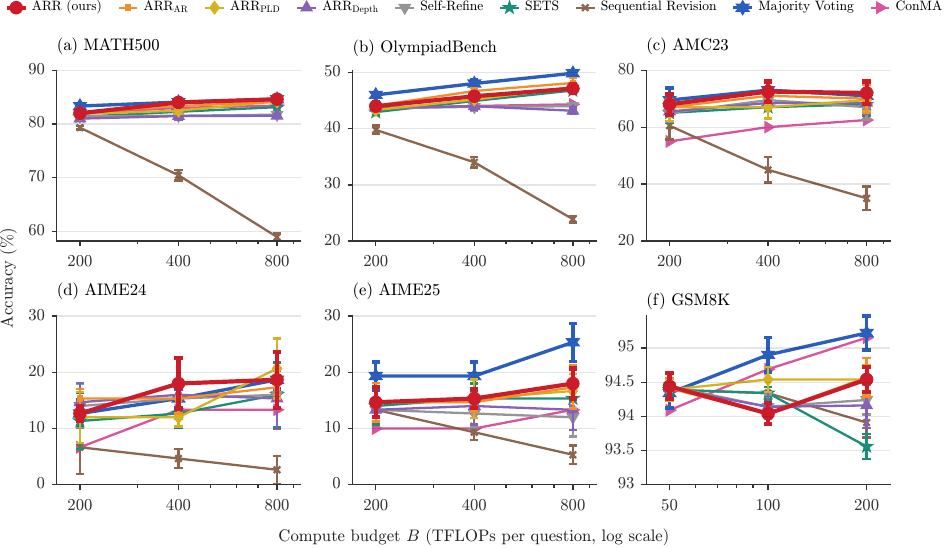}
\caption{Accuracy against the FLOP budget $B$ (Qwen2.5-Math-7B-Instruct). Error bars show the standard deviation over seeds.}
\label{fig:budget}
\end{figure*}

\begin{figure*}[t]
\centering
\includegraphics[width=1\textwidth]{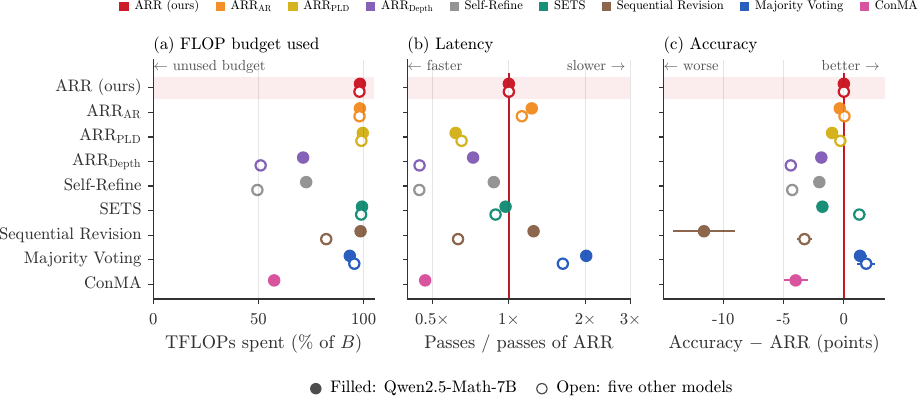}
\caption{Compute and accuracy over all settings: (a) TFLOPs used, (b) serial passes relative to ARR, and (c) accuracy difference from ARR.}
\label{fig:compute}
\end{figure*}

\textbf{Cost of a position.} The accounting gives a FLOP cost to each processed position.
A position that attends to $c$ cached keys costs
\begin{equation}
\kappa(c) = 2\,|\theta| + 4\, n_{\mathrm{layer}}\, d_{\mathrm{attn}}\, c .
\label{eq:flops}
\end{equation}
The first term is the cost of the matrix multiplications, and $|\theta|$ is the number of matrix-multiplication parameters.
The second term is the cost of attention, where $n_{\mathrm{layer}}$ is the number of layers and $d_{\mathrm{attn}}$ is the attention width.

\textbf{Charging rules.} The accounting charges every position in $\mathcal{P}_\pi(x)$ (Section~\ref{sec:budget}) before the pass that processes it, and it refunds nothing.
If a draft block does not fit in the remaining budget, the block is shortened.
A new root is created only if the remaining budget, minus the worst-case cost of the calls in progress, covers a complete sample.
For hybrid linear-attention models, the positions that are fed again after a rejected block (Appendix~\ref{app:impl}) are charged as recomputation.

\textbf{Budgets.}\label{app:budgets} We convert a budget $B$ to a number of samples.
A sample with a prompt of $P$ tokens and an output of $T$ tokens processes $L=P+T$ positions, and position $i$ attends to $i$ keys.
The sum of Equation~(\ref{eq:flops}) over these positions gives the cost of one sample,
\begin{equation}
C(L) \approx 2\,|\theta|\,L + 2\, n_{\mathrm{layer}}\, d_{\mathrm{attn}}\, L^2 .
\label{eq:sample-cost}
\end{equation}
A budget $B$ thus buys approximately $K = B/C(L)$ samples, and at least $K_{\min} = B/C(P+\Tmax)$ complete samples.
For Qwen2.5-Math-7B-Instruct, $|\theta|\approx7.1$B, $n_{\mathrm{layer}}{=}28$, and $d_{\mathrm{attn}}{=}3{,}584$.
With $P\approx200$ and $\Tmax{=}1{,}536$, Equation~(\ref{eq:sample-cost}) gives $C\approx25$ TFLOPs, and the attention term is less than 3\% of this cost.
Thus, $B\in\{200,400,800\}$ TFLOPs gives $K_{\min}\approx8$, 16, and 32.
Most solutions are shorter than $\Tmax$: on MATH500 at $B{=}800$, \SC{} collects approximately 97 samples, i.e., approximately 8 TFLOPs for each sample (Table~\ref{tab:cost}).
GSM8K solutions are shorter again, so GSM8K uses one quarter of these budgets, $B\in\{50,100,200\}$ TFLOPs.
For a model $m$ with $\Tmax{=}4{,}096$, we scale the budget so that $K_{\min}$ stays the same:
\begin{equation}
B_m \approx B \cdot \frac{|\theta_m|}{|\theta_{\mathrm{7B}}|} \cdot \frac{4{,}096}{1{,}536},
\label{eq:budget-scale}
\end{equation}
where $\theta_{\mathrm{7B}}$ are the parameters of Qwen2.5-Math-7B-Instruct.
This gives $\{130,260,520\}$ for Qwen3.5-2B, $\{350,700,1400\}$ for Gemma-4, and $\{650,1300,2600\}$ for Qwen3.5-9B.
A latency-matched variant limits the serial passes per question to $\{2{,}000, 4{,}000, 8{,}000\}$.

\section{Compared Methods}
\label{app:methods}

\begin{figure*}[t]
\centering
\includegraphics[width=1\textwidth]{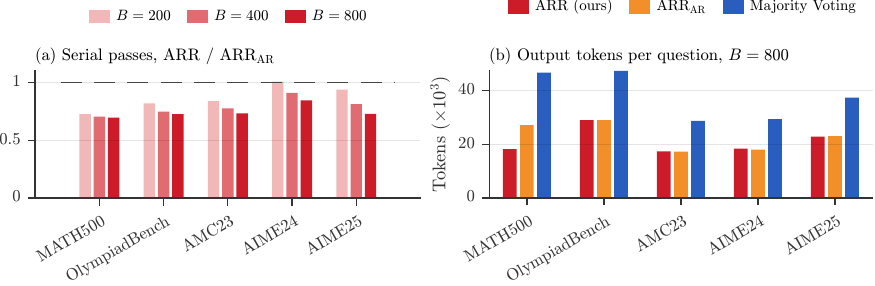}
\caption{Cost of revision for Qwen2.5-Math-7B-Instruct. (a) Ratio of the median serial passes of ARR to those of \ARRar{}. Replay gives a larger decrease at larger budgets. (b) Output tokens for each question at $B{=}800$.}
\label{fig:replay}
\end{figure*}

\begin{figure*}[t]
\centering
\includegraphics[width=0.9\textwidth]{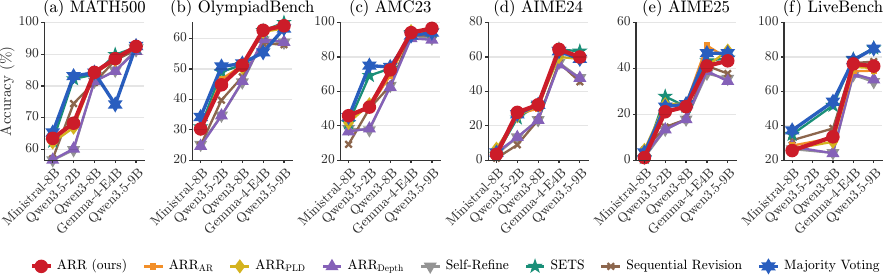}
\caption{Accuracy (\%) of every method on the other models at their largest budget ($B{=}800$ for Ministral-8B and Qwen3-8B, 520 for Qwen3.5-2B, 1,400 for Gemma-4-E4B, and 2,600 for Qwen3.5-9B).}
\label{fig:models}
\end{figure*}

Table~\ref{tab:methods} summarizes the methods of Section~\ref{sec:exp-setup}.
This section gives only the details that the main text does not give.

\textbf{Prompt lookup.}
At each step, \ARRpld{} matches the last $n\le 3$ tokens against the prompt and the output, drafts the next 10 tokens, verifies them in one pass, and repeats the lookup after every mismatch; because $y$ is part of $\rho(x,y)$, it can realign to $y$ after a change.

\textbf{Why exact variants differ in accuracy.}
The exactness of replay (Section~\ref{sec:exact}) concerns a single revision, whereas a full run under a shared FLOP budget can diverge for three reasons.
\textit{(i) Cost.} The accounting charges every verified draft position, including rejected ones (Section~\ref{sec:budget}), so the same revision costs a different number of FLOPs under each decoder, and the budget ends at a different point.
\textit{(ii) Scheduling.} The decoders finish revisions at different speeds, so the concurrent roots of a question consume the budget in a different order.
This changes the reserve for calls in progress, which root receives the last affordable new sample, and which revision the budget truncates; the variants therefore create different numbers of roots (16.2, 15.9, and 13.3 per MATH500 question at $B{=}800$, Table~\ref{tab:cost}).
\textit{(iii) Numerics.} One near-tie broken differently by a batched and a serial pass flips a token and changes the rest of the trajectory of that root, including whether it reaches a fixed point.
Accuracy differences among the three variants therefore reflect these run-level effects, which are largest on benchmarks with 30-40 questions, not different revision operators.

\textbf{ConMA.} We implement ConMA \citep{wang-etal-2026-conma} with the hyperparameters of the paper (8 trajectories per round, weights 0.3 and 0.7 for the full-sequence and tail confidence, a tail of 128 tokens, a consensus weight of 0.5, a kernel threshold of 0.75, and an expansion trigger of 0.5) and its verbatim prompts. A trajectory gets a confidence score from the geometric mean of its token probabilities over the full sequence and over its tail; answer groups are ranked by consensus and mean score, and the kernel is the smallest set of top groups with a cumulative probability of at least the kernel threshold. If the kernel is large, ConMA samples new trajectories from a critical-review prompt; otherwise it asks a multiple-choice question over the kernel. It stops when one answer remains or when the budget cannot fund another round, and it shares the FLOP accounting and the answer extractor of the other methods. 

\textbf{Stopping of the other baselines.}
\ARRdepth{} stops when all $N$ roots are retired or truncated and leaves the rest of the budget unused.
\FDR{} also revises roots that are already fixed points.
\SeqC{} continues its roots round-robin and never retires or creates a root.

\begin{table*}[t]
\centering
\caption{Compared methods. The rows from \ARRar{} to \VC{} change one component of ARR each.}
\label{tab:methods}
\vspace{2pt}
\small
\setlength{\tabcolsep}{4pt}
\resizebox{\textwidth}{!}{%
\begin{tabular}{llllll}
\toprule
Method & Update of a root & Decoding of the update & Explicit verifier & A root stops when & Slot of a stopped root \\
\midrule
\rowcolor{oursrow}\textbf{ARR} & $F_x(y)$ & replay ($W{=}32$) & no & $F_x(y)=y$ or $F_x(y)\notin\Ycal$ & new root \\
\ARRar & $F_x(y)$ & autoregressive greedy & no & same as ARR & new root \\
\ARRpld & $F_x(y)$ & prompt lookup & no & same as ARR & new root \\
\ARRdepth & $F_x(y)$ & replay & no & same as ARR & not reused \\
\FDR & $F_x(y)$ & autoregressive greedy & no & after $D{=}4$ revisions & not reused \\
\VC{} & $F_x^{\mathrm{vc}}(y)$ & autoregressive greedy & yes & $F_x^{\mathrm{vc}}(y)=y$ or truncated & new root \\
\SeqC & sample from $\rho_{\mathrm{cont}}(x,y,k)$ ($\tau{=}0.7$) & autoregressive sampling & no & budget ends & - \\
\SC & none (new sample) & autoregressive sampling & no & at once & new sample \\
ConMA & none (groups of samples) & autoregressive sampling & confidence score & one answer remains & - \\
\bottomrule
\end{tabular}}

\end{table*}

\section{Prompt Templates}
\label{app:prompts}

\begin{figure*}[t]
\centering
\includegraphics[width=0.9\textwidth]{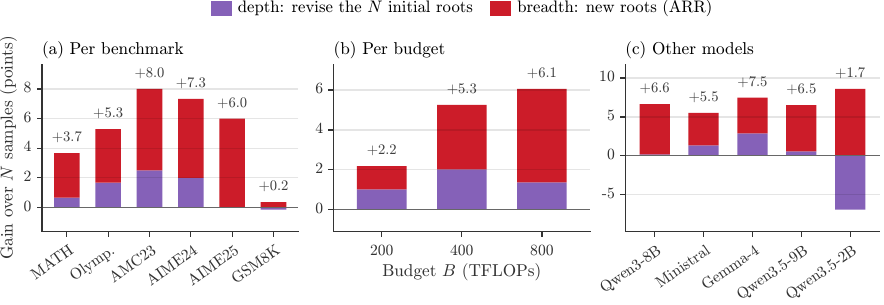}
\caption{Accuracy gains from revision depth and replacement breadth across benchmarks, budgets, and models.}
\label{fig:depthacc}
\end{figure*}

\begin{figure*}[t]
\centering
\includegraphics[width=0.9\textwidth]{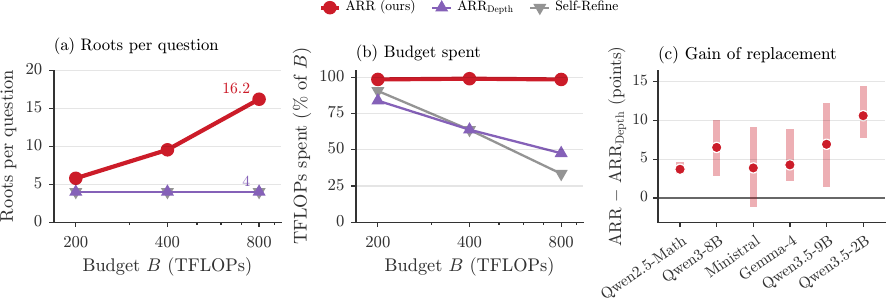}
\caption{Depth-to-breadth reallocation under fixed-point replacement across budgets and models.}
\label{fig:depth}
\end{figure*}

\begin{table*}[t]
\centering
\caption{Stopping rules with the replacement of ARR for Qwen2.5-Math-7B-Instruct. Answer: retire a root when its extracted answer does not change; Once: retire a root after one revision. $\Delta$: Answer $-$ ARR with a 95\% paired CI. Roots: roots per question of ARR / Answer. Passes: serial passes of Answer / ARR. All columns use the seeds of the stopping-rule runs, so $\Delta$ is paired. On MATH500, OlympiadBench, and GSM8K, these are a subset of the main-run seeds, so ARR and \SC{} differ from the all-seed means of Figure~\ref{fig:budget} (e.g., 84.0\% vs.\ 84.6\% on MATH500 at $B{=}800$). In all tables, \colorbox{bestcell}{\textbf{blue}} marks the best value and \colorbox{secondcell}{red} the second-best value.}
\label{tab:stop}
\vspace{2pt}
\small
\resizebox{\textwidth}{!}{%
\begin{tabular}{lr cccc c c c}
\toprule
Benchmark & $B$ & ARR & Answer & Once & \SC & $\Delta$ & Roots & Passes \\
\midrule
\multirow{3}{*}{MATH500} & 200 & 81.6 & 81.6 & \second{81.8} & \best{83.2} & $+0.0$ $[-1.1,+1.2]$ & 5.8 / 9.3 & 1.20 \\
 & 400 & \best{83.8} & 83.0 & \second{83.4} & 83.2 & $-0.8$ $[-2.4,+0.8]$ & 9.5 / 19.1 & 1.19 \\
 & 800 & 84.0 & \best{84.6} & \second{84.4} & 84.0 & $+0.6$ $[-0.8,+1.8]$ & 16.1 / 38.5 & 1.16 \\
\cmidrule(lr){1-9}
\multirow{3}{*}{OlympiadBench} & 200 & 44.0 & \second{45.0} & 44.0 & \best{46.4} & $+1.0$ $[-0.7,+2.7]$ & 5.2 / 7.0 & 1.14 \\
 & 400 & 45.7 & 46.9 & \second{47.6} & \best{48.1} & $+1.2$ $[-1.2,+3.6]$ & 8.2 / 14.0 & 1.19 \\
 & 800 & 47.6 & \best{49.9} & \second{48.2} & \best{49.9} & $+2.2$ $[+0.3,+4.2]$ & 13.3 / 27.8 & 1.18 \\
\cmidrule(lr){1-9}
\multirow{3}{*}{AMC23} & 200 & \second{68.0} & \second{68.0} & 67.5 & \best{69.5} & $+0.0$ $[-5.5,+5.5]$ & 5.0 / 7.0 & 1.15 \\
 & 400 & \second{72.5} & 68.0 & \second{72.5} & \best{73.0} & $-4.5$ $[-9.5,-0.5]$ & 7.7 / 14.2 & 1.18 \\
 & 800 & \second{72.0} & \best{73.0} & 71.5 & 71.0 & $+1.0$ $[-5.0,+7.5]$ & 12.8 / 28.3 & 1.23 \\
\cmidrule(lr){1-9}
\multirow{3}{*}{AIME24} & 200 & 12.7 & \second{14.0} & \best{16.7} & 12.7 & $+1.3$ $[-1.3,+4.7]$ & 5.5 / 6.6 & 1.06 \\
 & 400 & \best{18.0} & 14.0 & \second{17.3} & 15.3 & $-4.0$ $[-10.0,+0.7]$ & 8.1 / 12.9 & 1.06 \\
 & 800 & \best{18.7} & \second{17.3} & \second{17.3} & \best{18.7} & $-1.3$ $[-4.7,+2.0]$ & 13.1 / 24.9 & 1.09 \\
\cmidrule(lr){1-9}
\multirow{3}{*}{AIME25} & 200 & 14.7 & 17.3 & \second{18.0} & \best{19.3} & $+2.7$ $[-1.3,+6.7]$ & 5.4 / 6.0 & 1.08 \\
 & 400 & 15.3 & \best{22.0} & \second{20.0} & 19.3 & $+6.7$ $[+2.0,+11.3]$ & 7.7 / 11.5 & 1.13 \\
 & 800 & 18.0 & \second{22.0} & 21.3 & \best{25.3} & $+4.0$ $[-2.0,+10.7]$ & 12.5 / 22.6 & 1.14 \\
\cmidrule(lr){1-9}
\multirow{3}{*}{GSM8K} & 50 & \second{94.2} & \best{94.3} & \best{94.3} & \best{94.3} & $+0.1$ $[-0.5,+0.7]$ & 4.0 / 4.0 & 0.98 \\
 & 100 & 94.0 & 94.3 & \second{94.5} & \best{94.6} & $+0.3$ $[-0.4,+1.0]$ & 4.2 / 6.3 & 1.37 \\
 & 200 & 94.5 & \second{95.3} & \best{95.5} & 95.2 & $+0.8$ $[+0.1,+1.5]$ & 5.9 / 13.8 & 1.29 \\
\bottomrule
\end{tabular}}
\end{table*}

Italic text in braces marks an inserted field, such as the question or the current solution, and bold labels mark the system and user turns.
For LiveBench, the answer line uses the format of each task instead of \verb|\boxed{}|.

\begin{summarybox}{System prompt of all solve and revision calls}
Please reason step by step, and put your final answer within \ansfmt{\textbackslash boxed\{\}}.
\end{summarybox}

\begin{problembox}{Solve prompt $\sigma(x)$ (all methods)}
\ph{question}
\end{problembox}

\begin{correctionbox}{Revision prompt $\rho(x,y)$ (ARR, \ARRar, \ARRpld, \ARRdepth, \FDR)}
Check the proposed solution against the question. Return a complete corrected solution. Correct any reasoning or calculation errors, including their downstream consequences. \key{Preserve wording that does not need correction.} \key{If no correction is needed, reproduce the proposed solution exactly.} Output only the solution.

\smallskip
\role{Question} \ph{question}\\
\role{Proposed solution} \ph{solution}
\end{correctionbox}

\begin{critiquebox}{Critique prompt $\rho_{\mathrm{crit}}(x,y)$ (\VC, step 1)}
\role{System} You are a careful mathematical critic. You will be given a PROBLEM and a SOLUTION TRACE. Your job: identify any errors in the trace - logical mistakes, wrong calculations, incorrect assumptions, or faulty reasoning. If the trace is correct, say so explicitly. Be concise and specific. Do NOT solve the problem yourself.

\smallskip
\role{User} Review the following SOLUTION TRACE for the PROBLEM below. Identify any errors: wrong calculations, incorrect assumptions, logical mistakes, or missing steps. If no errors exist, state: `The solution appears correct.' Be specific - point to the exact step where an error occurs. Do NOT re-solve the problem.

\smallskip
PROBLEM \ph{question}\\
SOLUTION TRACE \ph{solution}\\
Extracted answer: \ph{answer}\\
CRITIQUE (list errors specifically, or confirm correctness):
\end{critiquebox}

\begin{critiquebox}{Correction prompt $\rho_{\mathrm{cor}}(x,y,\cdot)$ (\VC, step 2)}
\role{System} You are a careful mathematical editor. You will be given a PROBLEM, a SOLUTION TRACE, and a CRITIQUE. Your job: produce a corrected solution that fixes all errors identified in the critique while keeping the correct parts of the original trace. Use the same numbered-step format. End your response with exactly one line: \ansfmt{Final answer: \textbackslash boxed\{<answer>\}}

\smallskip
\role{User} You are given a PROBLEM, a SOLUTION TRACE, and a CRITIQUE of that trace. Produce a corrected solution: fix every error identified in the critique, keep the numbered-step format, preserve correct parts of the original trace, do NOT introduce new errors, end with exactly one line: \ansfmt{Final answer: \textbackslash boxed\{<answer>\}}

\smallskip
PROBLEM \ph{question}\\
ORIGINAL SOLUTION TRACE \ph{solution}\\
CRITIQUE \ph{critique}\\
CORRECTED SOLUTION (numbered steps, ending with \ansfmt{Final answer: \textbackslash boxed\{<answer>\}}):
\end{critiquebox}

\begin{reasonbox}{Continuation prompt $\rho_{\mathrm{cont}}(x,y,k)$ (\SeqC)}
\role{System} You are a precise mathematical reasoner. Solve problems using clearly numbered steps. Show every calculation inside each step. End your response with exactly one line: \ansfmt{Final answer: \textbackslash boxed\{<answer>\}}

\smallskip
\role{User} \key{You are at refinement depth} \ph{k} \key{of} \ph{D}. Below is the ORIGINAL PROBLEM and the CORRECTED REASONING from the previous depth. Your task:
\begin{enumerate*}[label=\arabic*.]
\item Review the corrected reasoning carefully.
\item Continue from Step \ph{n}: - extend or refine the solution, verifying each claim.
\item If the prior reasoning already reaches a correct answer with no errors, confirm it concisely. Do NOT change the answer unless you can identify a specific concrete error.
\item End with exactly one line: \ansfmt{Final answer: \textbackslash boxed\{<answer>\}}
\end{enumerate*}

\smallskip
PROBLEM \ph{question}\\
CORRECTED REASONING FROM DEPTH \ph{k-1} \ph{solution}
\end{reasonbox}

\section{A Worked Example of Replay and Replacement}
\label{app:example}

\begin{figure*}[t]
\centering
\includegraphics[width=0.9\textwidth]{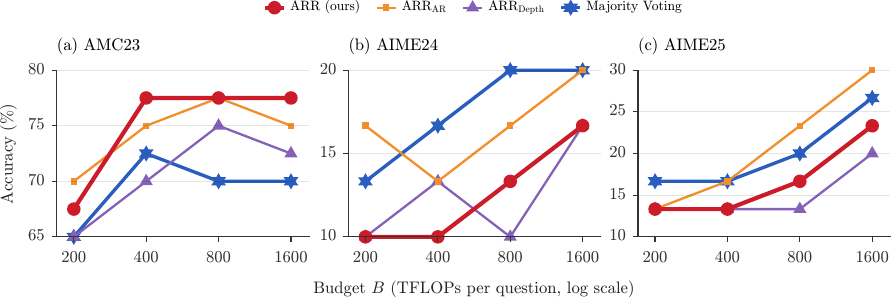}
\caption{Accuracy up to $B{=}1{,}600$.}
\label{fig:abl-budget}
\end{figure*}

\begin{figure*}[t]
\centering
\includegraphics[width=1\textwidth]{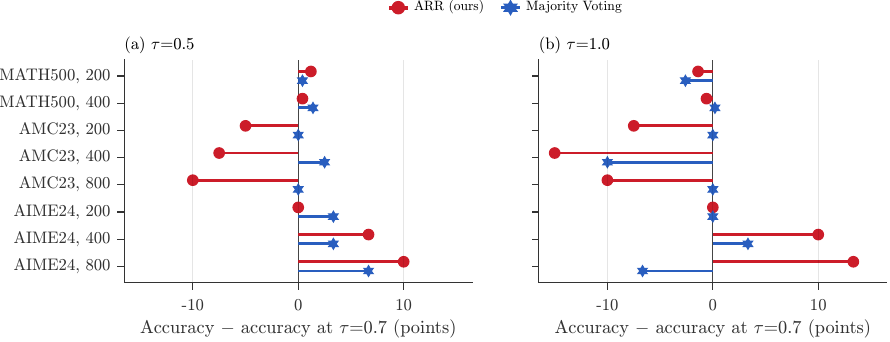}
\caption{Change of accuracy from the default $\tau{=}0.7$ at (a) $\tau{=}0.5$ and (b) $\tau{=}1.0$, for each benchmark and budget $B$.}
\label{fig:abl-temp}
\end{figure*}

We trace one question from the logs of Qwen2.5-Math-7B-Instruct on MATH500 (problem 358, $B{=}800$ TFLOPs).

\begin{problembox}{Problem (MATH500, no.\ 358)}
Determine if the graph of the equation below is a parabola, circle, ellipse, hyperbola, point, line, two lines, or empty.
\par\smallskip
\centerline{$\left(\frac{x}{2}-3\right)^2+y^2=10$}
\end{problembox}

The correct answer is \emph{ellipse}: the equation equals $(x-6)^2+4y^2=40$, in which the coefficients of $(x-6)^2$ and $y^2$ differ.
Many samples answer \emph{circle}, because they read the left side as a circle in the substituted variable $x/2$.

\textbf{Outcome.} Table~\ref{tab:example-outcome} compares the three methods on this question.
\SC{} and \VC{} return the frequent incorrect answer.
ARR returns the correct answer with 40\% fewer serial passes than \SC{}.

\begin{table}[tbp]
\centering
\footnotesize
\setlength{\tabcolsep}{4pt}
\resizebox{\columnwidth}{!}{%
\begin{tabular}{lllr}
\toprule
Method & Answer & Votes & Serial passes \\
\midrule
\SC{} & circle \xmark & 34 vs.\ 27 (61 samples) & 13{,}049 \\
\VC{} & circle \xmark & - & - \\
\rowcolor{oursrow}\textbf{ARR} & \textbf{ellipse} \cmark & 13 vs.\ 5 & 7{,}877 \\
\bottomrule
\end{tabular}}
\caption{Outcome on MATH500 problem 358 at $B{=}800$ TFLOPs. Votes give the winning answer against the runner-up. Serial passes are counted on the critical path.}
\label{tab:example-outcome}
\end{table}

\textbf{Trajectory of one root.} Table~\ref{tab:example} follows root~0 of ARR through one sample and four revisions.
Revisions 1 and 2 rewrite most of the solution but keep the answer \emph{circle}.
Revision 3 changes the answer to \emph{ellipse}, and revision 4 reproduces revision 3 exactly.

\begin{table}[tbp]
\centering
\footnotesize
\setlength{\tabcolsep}{2.2pt}
\begin{tabular}{lccccc}
\toprule
 & Sample & Rev.~1 & Rev.~2 & Rev.~3 & Rev.~4 \\
\midrule
Draft     & -     & 428    & 76     & 667    & 620 \\
Accepted  & -     & 0      & 0      & 602    & 620 \\
Discarded & -     & 0      & 0      & 6      & 0   \\
Decoded   & 428    & 75     & 666    & 17     & 0   \\
Passes    & 428    & 76     & 667    & 37     & 21  \\
Answer    & circle & circle & circle & \textbf{ellipse} & ellipse (FP) \\
\bottomrule
\end{tabular}
\caption{Trajectory of root~0. Accepted and discarded denote replay-verified draft tokens; decoded denotes greedy single-token steps, and FP denotes a fixed point.}
\label{tab:example}
\end{table}

\textbf{Revision 3: replay keeps the derivation and rewrites the end.} The draft of revision 3 has 667 tokens.
Its first 602 tokens derive $(x-6)^2+4y^2=40$ and state that this equation is an ellipse, but its last 65 tokens return to the circle form and box \emph{circle}.
Replay accepts the 602 tokens in 19 blocks, discards 6 positions of the failed block, and decodes a new 17-token end.
The call takes 37 serial passes, whereas autoregressive decoding takes 621 passes for the same output.
The three boxes below show the parts of this revision in order.
The logs store the token counts and the answer of each call, but not the full text of intermediate drafts, so the second box describes the discarded end.

\begin{reasonbox}{Accepted prefix of revision 3 (602 tokens, 19 blocks, verified in parallel)}
\ldots\ If we compare $(x-6)^2+4y^2=40$ to $(x-h)^2+(y-k)^2=r^2$, we see that the coefficients of $(x-6)^2$ and $y^2$ are not the same.
Therefore, the equation $(x-6)^2+y^2=10$ \key{does not represent a circle}.

\smallskip
The correct form of the equation is $(x-6)^2+4y^2=40$, \key{which represents an ellipse}.
\end{reasonbox}

\begin{incorrectbox}{Discarded end of the draft (65 tokens)}
\textit{Returns to the circle form and ends with the boxed answer} \textbf{circle}.
\end{incorrectbox}

\begin{correctionbox}{New end of revision 3 (17 tokens, greedy decoding)}
Therefore, the correct answer is: $\boxed{\textbf{ellipse}}$.
\end{correctionbox}

\textbf{Revision 4: a fixed point.} Revision 4 reproduces all 620 tokens, including the terminating $\eos$, in 21 serial passes.
Thus, root~0 is a fixed point: ARR keeps \emph{ellipse} as its vote and gives the slot to a new root.

\begin{observationbox}
\begin{itemize}[leftmargin=*,itemsep=2pt,topsep=0pt]
\item \SC{} counts each sample once, and thus a frequent shared error wins the vote.
\item ARR revises each sample until the revision is stable, and thus a root with the correct result in its derivation can correct its conclusion.
\item Revisions 1 and 2 do not change the answer, so a fixed schedule of two revisions keeps the incorrect answer.
\item ARR continues because the text still changes, and it stops as soon as the text stops changing.
\end{itemize}
\end{observationbox}

\section{Implementation Details}
\label{app:impl}

\begin{figure*}[t]
\centering
\includegraphics[width=1\textwidth]{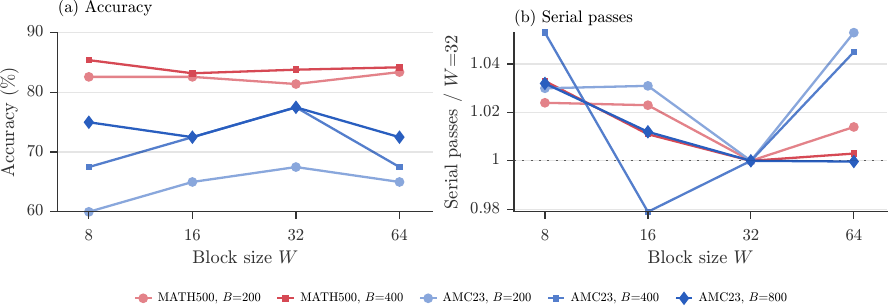}
\caption{(a) Accuracy and (b) serial passes relative to the default $W{=}32$, for each benchmark and budget.}
\label{fig:abl-window}
\end{figure*}

\begin{figure*}[t]
\centering
\includegraphics[width=1\textwidth]{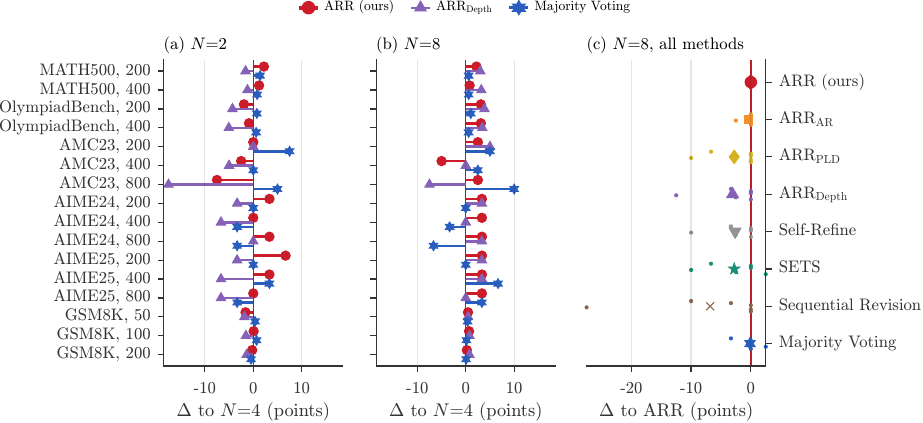}
\caption{Change of accuracy from the default $N{=}4$ at (a) $N{=}2$ and (b) $N{=}8$, for each benchmark and budget. (c) Accuracy relative to ARR ($N{=}8$) on AMC23 and AIME24; dots denote individual settings, large markers denote means.}
\label{fig:abl-width}
\end{figure*}

\textbf{Decoder.} All methods run in one decoder on HuggingFace Transformers, with continuous batching, a static key-value cache for each slot, and access to the logits.
Replay blocks, prompt-lookup blocks, and single-token steps run in separate batched passes.
The shared job queue (Appendix~\ref{app:algorithms}) is first-in-first-out, so a new root never overtakes an unfinished root.
The key-value cache of an accepted prefix is reused only within its call, so each revision prefills $\rho(x,y)$ again.

\textbf{Architectures.} The decoder detects the layer types of a model with one forward pass.
For hybrid linear-attention models (Qwen3.5), it keeps a recurrent state for each slot.
It verifies a draft block on a copy of this state and feeds the accepted prefix again after a rejection.
Each run uses one A100 or H100 GPU.

\textbf{Output equivalence of replay and greedy decoding.} Let $g=F_x(y)$ be the output of autoregressive greedy decoding on $u=\rho(x,y)$.
Replay and greedy decoding share the prefill of $u$, so their first output token is the same token $g_1$.
While the committed output is on the draft, $z=y_{1:t}=g_{1:t}$, replay runs one causal pass on $(z_t,y_{t+1},\dots,y_{t+m})$.
The causal mask restricts the logits at block position $j$ to $u$ and $y_{1:t+j-1}$, so $\hat z_{t+1}$ is the greedy token after $g_{1:t}$, which is $g_{t+1}$.
If this token equals the draft token $y_{t+1}$, the conditioning prefix of the next block position is again a prefix of $g$, and the same argument holds at every later position of the block.
Thus, the accepted span $y_{t+1:t+\lambda}$ and the appended token $\hat z_{t+\lambda+1}$ are exactly the next $\lambda{+}1$ tokens of $g$.
The positions after the first disagreement condition on a draft token that $g$ does not contain, so the decoder discards their key-value entries before the next pass, and no rejected position affects a later prediction.
After the output leaves the draft, replay continues with single-token greedy steps from a prefix of $g$ and thus produces the rest of $g$.
Both decoders stop at the same $\eos$ or at the same $\Tmax$, and an accepted $\eos$ inside a block ends the call at once.
For $W{=}1$, each block has $m{=}0$ draft tokens, and replay reduces to greedy decoding.
A fixed point is also stable under further revision: because $\rho$ contains no round index or history, $F_x$ maps each $y$ to one output, so $F_x(y)=y$ gives $F_x^{\,k}(y)=F_x\big(F_x^{\,k-1}(y)\big)=y$ for every $k\ge 1$.
This equivalence holds only if a batched pass and a single-token pass select the same argmax; the verification tests below check this condition on our backend.

\begin{table}[tbp]
\centering
\caption{Mechanism on MATH500 at $B{=}800$ TFLOPs with Qwen2.5-Math-7B-Instruct. TF: TFLOPs per question; Roots: roots per question; FP: fixed-point rate; W$\to$C, C$\to$W: answer changes (Appendix~\ref{app:impl}).}
\label{tab:cost}
\vspace{2pt}
\footnotesize
\resizebox{\columnwidth}{!}{%
\begin{tabular}{lrrrrr}
\toprule
Method & TF & Roots & FP & W$\to$C & C$\to$W \\
\midrule
\rowcolor{oursrow}\textbf{\ARR{}} & 789 & 16.2 & .23 & 663 & 357 \\
\ARRar    & 788 & 15.9 & .22 & 666 & 374 \\
\ARRpld   & 797 & 13.3 & .30 & 463 & 230 \\
\ARRdepth & 381 & 4.0  & .08 & 300 & 234 \\
\FDR      & 268 & 4.0  & .53 & 173 & 96  \\
\VC       & 793 & 14.8 & .69 & 244 & 258 \\
\SeqC     & 793 & 4.0  & -  & 218 & 278 \\
\SC       & 781 & 96.6 & -  & 0   & 0   \\
ConMA & 185 & 10.0 & - & - & - \\
\bottomrule
\end{tabular}}
\end{table}

\textbf{Mechanism statistics.} Table~\ref{tab:cost} reports five statistics per method, computed over all questions and seeds of a cell. \textit{TF} is the mean number of TFLOPs that the FLOP accounting of Section~\ref{sec:budget} charges per question, including prefills, discarded draft positions, and truncated calls. \textit{Roots} is the mean number of roots $|\mathcal{R}|$ created per question: the $N{=}4$ initial roots plus every new root $r'$; for \SC{} it is the number of samples, and methods without replacement stay at $N$. \textit{FP} is the fixed-point rate, the fraction of complete revisions ($z\in\Ycal$) that reproduce their input exactly, $z=y$; for \VC{} the test is $F_x^{\mathrm{vc}}(y)=y$, and it is undefined for \SeqC{}, whose update is sampled, and for \SC{}, which never revises. \textit{W$\to$C} and \textit{C$\to$W} count complete updates whose extracted answer $\phi$ changes from incorrect to correct and from correct to incorrect with respect to $a^\star(x)$, summed over all questions and seeds; these labels are used only for evaluation and never by a method. Updates that keep the correctness of the answer are not counted, so W$\to$C$-$C$\to$W is the net number of answers that revision repairs. We also record the output tokens per question and the accepted-prefix fraction $\ell/|y|$ of complete revisions.


\section{Additional Results}
\label{app:results}

\begin{figure*}[t]
\centering
\includegraphics[width=1\textwidth]{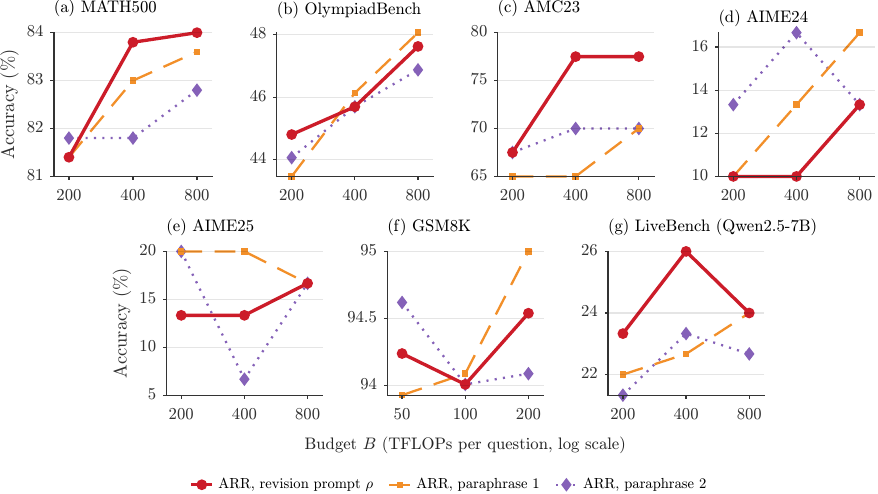}
\caption{Accuracy of ARR with the revision prompt $\rho$ and two paraphrases; (g) uses Qwen2.5-7B-Instruct.}
\label{fig:abl-prompts}
\end{figure*}

\begin{figure*}[t]
\centering
\includegraphics[width=1\textwidth]{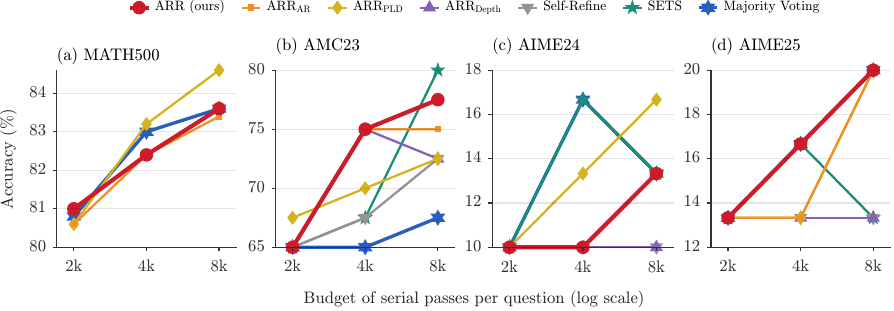}
\caption{Accuracy against a budget of serial passes per question. \SeqC{} is omitted for scale.}
\label{fig:abl-passes}
\end{figure*}

\begin{figure*}[t]
\centering
\includegraphics[width=0.62\textwidth]{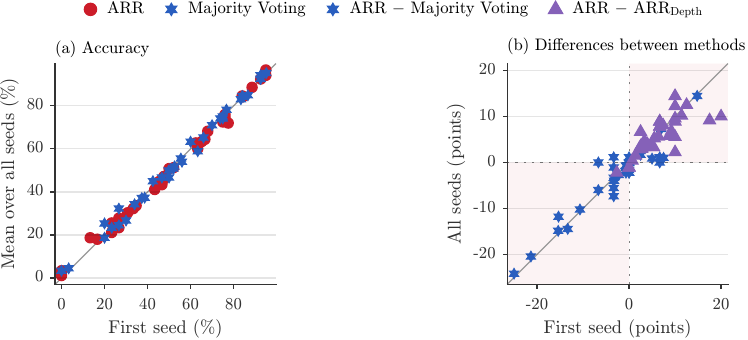}
\caption{First seed against the mean over all seeds at the largest budget. (a)~Accuracy of ARR and \SC{}. (b)~Accuracy differences between methods.}
\label{fig:seeds}
\end{figure*}

\textbf{Equal-FLOP view.} Figure~\ref{fig:budget} gives accuracy against the FLOP budget, and Figure~\ref{fig:compute} gives the compute that each method uses.
Six methods spend 93-100\% of $B$ on Qwen2.5-Math-7B-Instruct, so their FLOPs are equal by design.
\ARRdepth{} stops when its roots get to fixed points, and \FDR{} stops after $D{=}4$ revisions; neither reuses the slots. They spend only 50-73\% of $B$ and lose 1.9-4.4 points.
ConMA stops when one answer remains and spends 57\% of $B$.

\textbf{Adaptive allocation.} ARR does not stop at a fixed point. It gives the remaining budget to a new root and thus spends 98\% of $B$.
On Qwen2.5-Math-7B-Instruct, ARR is the most accurate method that revises, and \SC{} is 1.3 points better on average over all budgets.
On the five other models at the largest budget, \VC{} and \SC{} are 1.7 and 1.5 points better on average (Figure~\ref{fig:models}, Section~\ref{sec:models}).
On the primary model and on the other models, \SC{} needs 2.0 and 1.6 times the serial passes of ARR, and \ARRar{} needs 1.2 and 1.1 times (Figure~\ref{fig:compute}(b)).
ARR stays within 0.3 points of \ARRar{}. Thus, ARR gives more accuracy for each serial pass at the FLOP cost of majority voting.

\begin{table}[t]
\centering
\caption{Accuracy (\%) of the stopping rules of Table~\ref{tab:stop} for Qwen3-8B.}
\label{tab:stop-qwen3}
\vspace{2pt}
\small
\begin{tabular}{lr ccc}
\toprule
Benchmark & $B$ & ARR & Answer & Once \\
\midrule
\multirow{3}{*}{AMC23} & 200 & 62.5 & 62.5 & 62.5 \\
 & 400 & \second{65.0} & \best{72.5} & \second{65.0} \\
 & 800 & \second{75.0} & 72.5 & \best{77.5} \\
\cmidrule(lr){1-5}
\multirow{3}{*}{AIME24} & 200 & \best{26.7} & \second{23.3} & 20.0 \\
 & 400 & \best{26.7} & \second{23.3} & \second{23.3} \\
 & 800 & \best{33.3} & 23.3 & \second{26.7} \\
\cmidrule(lr){1-5}
\multirow{3}{*}{AIME25} & 200 & 16.7 & 16.7 & 16.7 \\
 & 400 & \second{23.3} & \second{23.3} & \best{26.7} \\
 & 800 & \best{26.7} & \second{20.0} & \second{20.0} \\
\bottomrule
\end{tabular}
\end{table}

\textbf{Detailed values for Section~\ref{sec:results}.} On MATH500 at $B{=}800$, ARR and \SC{} both get 84.6\% (paired 95\% CI $[-1.2,+1.3]$).
ARR writes 18.2k output tokens for each question and collects 15.0 votes. \SC{} writes 46.7k tokens and collects 93.3 votes.
ConMA uses 23\% of $B$ (Table~\ref{tab:cost}) and 0.12 of the serial passes of ARR; averaged over all settings, it uses 57\% of $B$ and 0.47 of the serial passes.

\textbf{Wall-clock latency.} Table~\ref{tab:timing} gives the latency on a subset of MATH500 problems at $B{=}400$, so its accuracies are not comparable to Figure~\ref{fig:budget}. One A100 GPU runs one question at a time, and no other job uses the GPU.
The median latency for each question is 78.5\,s for ARR, 83.3\,s for \ARRar{}, 42.5\,s for \ARRpld{}, 176.5\,s for \SC{}, and 18.4\,s for ConMA.
All methods get 90.0\%, except ConMA, which gets 87.5\% and uses only 23\% of $B$.
A replay pass processes up to $W$ positions. Thus, ARR is only 6\% faster than \ARRar{} in wall-clock time, although it needs 30\% fewer serial passes.
\ARRpld{} needs the fewest serial passes and is the fastest method that revises.

\begin{table}[t]
\centering
\caption{Wall-clock latency on one A100 GPU (Qwen2.5-Math-7B-Instruct, a subset of MATH500 problems, $B{=}400$, one question at a time). Latency and serial passes are medians for each question; accuracy is on this subset.}
\label{tab:timing}
\vspace{2pt}
\small
\setlength{\tabcolsep}{4pt}
\begin{tabular}{lrrrr}
\toprule
Method & Latency (s) & Passes & TFLOPs & Acc. \\
\midrule
\rowcolor{oursrow}\textbf{ARR} & 78.5 & 2{,}748 & 398 & 90.0 \\
\ARRar & 83.3 & 3{,}926 & 398 & 90.0 \\
\ARRpld & \second{42.5} & \second{1{,}593} & 399 & 90.0 \\
\SC & 176.5 & 6{,}878 & 381 & 90.0 \\
ConMA & \best{18.4} & \best{644} & 93 & 87.5 \\
\bottomrule
\end{tabular}
\end{table}

\textbf{Detailed values for Section~\ref{sec:ablation}.} The roots of ARR increase with the budget: 5.8, 9.5, and 16.2 for each MATH500 question.
\ARRdepth{} uses 381 of 800 TFLOPs on MATH500, and \FDR{} uses 268 TFLOPs for 81.7\% (Figure~\ref{fig:depth}).
ARR is better than \ARRdepth{} by 3.1 points $[2.0,4.3]$ on MATH500 and by 4.0 points $[2.6,5.2]$ on OlympiadBench.
The depth gain stays at 1.0-2.0 points at all budgets (Figure~\ref{fig:depthacc}).
\VC{} is 1.5 points $[0.3,2.7]$ less accurate than ARR on MATH500, and 69\% of its updates are fixed points.
\SeqC{} decreases from 79.3\% to 59.0\% on MATH500 as the budget increases, and it is 25.6 points $[22.5,28.5]$ below ARR at $B{=}800$.
\ARRpld{} uses 0.53-0.58 of the serial passes of ARR at $B{=}800$.

\textbf{Detailed values for Section~\ref{sec:models}.}
Across mathematical reasoning settings for other models, ARR achieves mean gains of 6.4 and 6.3 points over \ARRdepth{} and \FDR{}, respectively.
\SC{} outperforms ARR in more settings than the reverse.
Excluding Qwen3.5-2B and the two Gemma-4 settings with repaired formatting, \VC{} and ARR perform within 0.1 points of each other.

\textbf{Ablations.} We change one setting of ARR at a time with Qwen2.5-Math-7B-Instruct, and we compare each run with the matched main run.
We fixed the default settings ($N{=}4$, $\tau{=}0.7$, $W{=}32$) before these runs, and all other experiments use them.
Table~\ref{tab:stop} gives the stopping rules of Figure~\ref{fig:regimes} at all budgets, and Table~\ref{tab:stop-qwen3} gives the same rules for Qwen3-8B.
Figures~\ref{fig:abl-temp}, \ref{fig:abl-window}, \ref{fig:abl-budget}, \ref{fig:abl-width}, \ref{fig:abl-prompts}, and~\ref{fig:abl-passes} give the sampling temperature, the block size $W$, the budget $B{=}1{,}600$, the width $N$, the paraphrases of $\rho$, and the latency-matched budgets.
Each ablation shows a benchmark and budget only if all compared settings and methods finished there. A result in a figure does not occur again in a table.

\textbf{Detailed values of the ablations.}
Across settings with complete results for $N\in\{2,4,8\}$, ARR with $N{=}8$ gains 1.9 points over $N{=}4$ while using 0.64 times as many serial passes.
At $N{=}8$, ARR is within 0.5 points of \SC{} while using 0.59 times as many serial passes.
A block size $W\in[8,64]$ changes the serial passes by at most 5\%.
At $B{=}1{,}600$, ARR achieves 39.2\%, \SC{} 38.9\%, and \ARRdepth{} 36.4\% accuracy on average across AMC23, AIME24, and AIME25.
The two paraphrases of $\rho$ change accuracy by at most 1.2 points on MATH500, OlympiadBench, and GSM8K.

\textbf{One seed and several seeds.} Figure~\ref{fig:seeds} compares first-seed accuracy with the mean across all seeds. First-seed accuracy differs from the all-seed mean by at most 5.5 points. The sign of ARR$-$\ARRdepth{} agrees between the first seed and the all-seed mean in 94\% of settings, compared with 71\% for ARR$-$\SC{}. Among the disagreements with \SC{}, 80\% involve an exact tie in one of the comparisons, and the all-seed mean differences remain within 2.2 points. Thus, single-seed comparisons with \SC{} are less reliable, while ARR's gains over \ARRdepth{} are more stable.

\section{Extended Related Work}
\label{app:related}

\begin{table*}[t]
\centering
\caption{Signals that end or redirect refinement in the closest methods. The signal of ARR is an exact property of a deterministic operator, not a verdict or a score. ARR gets the signal with parallel draft verification, which comes from prior work \citep{sun-etal-2021-instantaneous,pmlr-v202-leviathan23a,saxena2023pld}.}
\label{tab:positioning}
\vspace{2pt}
\footnotesize
\setlength{\tabcolsep}{5pt}
\resizebox{\textwidth}{!}{%
\begin{tabular}{lcccl}
\toprule
\multirow{2}{*}{Method} & Training- & No step & Released compute & \multirow{2}{*}{Signal that stops or redirects refinement} \\
 & free & structure & $\rightarrow$ new attempts & \\
\midrule
Self-Refine \citep{madaan2023selfrefineiterativerefinementselffeedback} & \cmark & \cmark & \xmark & stop word in self-feedback, or iteration limit \\
SETS \citep{chen2025setsleveragingselfverificationselfcorrection} & \cmark & \cmark & \xmark & self-verification verdict ``correct'', or round limit \\
Thought-ICS \citep{pmlr-v306-samanta26a} & \cmark & \xmark & \xmark & self-localization of the first incorrect thought \\
Adaptive-Consistency \citep{aggarwal2023letssamplestepstep} & \cmark & \cmark & \xmark & agreement of the vote distribution (stops sampling) \\
ConMA \citep{wang-etal-2026-conma} & \cmark & \cmark & \cmark & token-probability confidence and answer frequency \\
DeepConf \citep{fu2025deepthinkconfidence} & \cmark & \cmark & \xmark & token-confidence threshold on traces \\
Compute-optimal \citep{snell2024scalingllmtesttimecompute} & \xmark & \cmark & \cmark & learned verifier and predicted difficulty \\
SCoRe \citep{kumar2024traininglanguagemodelsselfcorrect} & \xmark & \cmark & \xmark & fixed number of trained correction turns \\
\midrule
\rowcolor{oursrow}\textbf{ARR (ours)} & \cmark & \cmark & \cmark & exact fixed point $F_x(y)=y$, found during decoding \\
\bottomrule
\end{tabular}}
\end{table*}

\subsection{More draft-and-verify methods.} Medusa \citep{cai2024medusasimplellminference} drafts with extra heads, lookahead decoding \citep{fu2024breaksequentialdependencyllm} drafts with Jacobi iterations, and EAGLE \citep{li2025eaglespeculativesamplingrequires} drafts with an extra head on the hidden features.

\subsection{Sequential scaling and efficient reasoning.} Long reasoning traces frequently spend tokens on redundant reflection \citep{chen2025think23overthinkingo1like}, including those of reasoning models trained with reinforcement learning \citep{deepseek2025r1}.
Budget forcing extends or ends a reasoning trace to control its length \citep{muennighoff2025s1simpletesttimescaling}, and \citet{sui2025stopoverthinkingsurveyefficient} survey such efficient reasoning methods.
ARLCP \citep{yu2026stopunnecessaryreflectiontraining} trains models to reflect less with length-sensitive rewards.
ARR keeps the weights frozen and removes a different redundancy: the repeated generation of unchanged text across revision rounds.

\subsection{Fixed points in sequence models.} Deep equilibrium models compute fixed points of layers \citep{bai2019deepequilibriummodels}, and Jacobi decoding iterates a parallel greedy update to its fixed point \citep{santilli2023acceleratingtransformerinferencetranslation,pmlr-v235-kou24a}.
In these methods, the fixed point is the output.
In ARR, the fixed point is a stop signal for one solution in a population of solutions.

Inference servers reuse cached prefixes across requests \citep{kwon2023efficientmemorymanagementlarge,zheng2024sglangefficientexecutionstructured}, which is complementary to replay.

\end{document}